\RequirePackage{fix-cm}

\documentclass[twocolumn]{svjour3}          
\smartqed  
\usepackage{microtype}

\usepackage{graphicx}
\usepackage{amsmath}
\usepackage{amssymb}
\usepackage{physics}
\usepackage{xcolor}
\usepackage{booktabs}
\usepackage{multirow}

\usepackage{natbib}
\setcitestyle{aysep={}} 

\usepackage{setspace}
\usepackage{natbib}

\usepackage[ruled,vlined]{algorithm2e}

\usepackage{pifont}

\usepackage{bm}
\usepackage{caption}

\usepackage{wrapfig}
\usepackage{multirow}
\usepackage{breakcites}

\usepackage[pagebackref=true,breaklinks=true,letterpaper=true,colorlinks,bookmarks=false]{hyperref}
\hypersetup{citecolor = blue}

\newcommand{\mbf}[1]{\mathbf{#1}}
\newcommand{\mth}{\boldsymbol{\theta}}
\newcommand{\mbt}{\boldsymbol{\beta}}

\newcommand{\loss}{\mathcal{L}}
\newcommand{\mI}{\mathcal{I}}
\newcommand{\mS}{\mathcal{S}}
\newcommand{\mUV}{UV}

\usepackage{booktabs}

\usepackage{color}

\newcommand{\etal}{\textit{et al}.}
\newcommand{\ie}{\textit{i}.\textit{e}.}
\newcommand{\eg}{\textit{e}.\textit{g}.}

\journalname{International Journal of Computer Vision}

\begin{document}

\title{A Shape-Aware Retargeting Approach to Transfer Human Motion and Appearance in Monocular Videos}

\author{Thiago L. Gomes$^1$ \and Renato Martins$^2$ \and Jo\~ao Ferreira$^1$ \and Rafael Azevedo$^1$ \and Guilherme Torres$^1$ \and Erickson R. Nascimento$^1$
}
\authorrunning{Thiago Gomes~\etal} 

\institute{
	$^1$ \at Department of Computer Science -- Universidade Federal de Minas Gerais, Belo Horizonte, Brazil. \\
	\email{\{thiagoluange, joaoferreira, torres, erickson\}@dcc.ufmg.br, rafaelvieiraz@ufmg.br} 
	\and
	$^2$ \at INRIA, Sophia Antipolis, France. \\
	\email{renato.martins@inria.fr}    \\
	}

\date{Received: date / Accepted: date}

\maketitle

\begin{abstract}
Transferring human motion and appearance between videos of human actors remains one of the key challenges in Computer Vision. Despite the advances from recent image-to-image translation approaches, there are several transferring contexts where most end-to-end learning-based retargeting methods still perform poorly. Transferring human appearance from one actor to another is only ensured when a strict setup has been complied, which is generally built considering their training regime's specificities. In this work, we propose a shape-aware approach based on a hybrid image-based rendering technique that exhibits competitive visual retargeting quality compared to state-of-the-art neural rendering approaches. The formulation leverages the user body shape into the retargeting while considering physical constraints of the motion in 3D and the 2D image domain. We also present a new video retargeting benchmark dataset composed of different videos with annotated human motions to evaluate the task of synthesizing people's videos, which can be used as a common base to improve tracking the progress in the field. The dataset and its evaluation protocols are designed to evaluate retargeting methods in more general and challenging conditions. Our method is validated in several experiments, comprising publicly available videos of actors with different shapes, motion types, and camera setups. The dataset and retargeting code are publicly available to the community at: \url{https://www.verlab.dcc.ufmg.br/retargeting-motion}.
\end{abstract}

\section{Introduction}\label{sec:introduction}

Human image synthesis has seen fast progress from the realistic image-to-image translation of human characters, as milestones are being passed in a wide range of applications, including face synthesis, style transferring, and motion retargeting. Despite remarkable advances in synthesizing virtual moving actors with image-to-image translation approaches, creating plausible virtual actors from images of real actors still remains a key challenge. These approaches face the unique cognitive capability of humans for perceiving attributes of objects from images. Humans start learning early in their lives to recognize human forms and make sense of what emotions and meanings are being communicated by human movement. We are, by nature, specialists in the human form and movement analysis. Even small imperfections when synthesizing virtual actors might create a false appearance, especially body motion shaking effects when dealing with moving actors.

Many works in the Computer Vision and Computer Graphics communities have made great strides in capturing human geometry and motion through model-based and learning techniques. In particular, end-to-end learning approaches such as~\cite{2018-TOG-SFV},~\cite{kanazawaHMR18}, and~\cite{kolotouros2019spin} have achieved state of the art in capturing three-dimensional motion, shape, and appearance from videos and still images from real actors. Most recently, several methods have been proposed on body reenactment from source images~\citep{Lassner_GeneratingPeople,ZhaoWCLF17,ma2017pose,chan2018dance,Esser_2018_CVPR,lwb2019}. The ultimate goal of these methods is to create a video where the body of a target person is reenacted according to the motion extracted from the monocular video. The motion is estimated considering the set of poses of a source person. Despite the impressive results for several input conditions, there are instances where most of these methods perform poorly. For instance, the works of~\cite{chan2018dance} and~\cite{wang2018vid2vid}, only perform good reenacting of the appearance/style from one actor to another if a strict setup has complied, \eg, static backgrounds, a large set of motion data of the target person to train, and actors in the same distance from the camera~\citep{tewari2020cgf}. Furthermore, it is hard to gauge progress from these results in the field of retargeting, as most works only report the performance of their algorithms in their own set of images which, in general, is built considering the specificities in the training regime of their approaches.

In this paper, we tackle the problem of not only body reenactment but transferring a target actor from a video sequence to a new video. In the synthesized new video, the target must perform a similar motion to a source actor considering his/her physical motion constraints and the video's background where the source actor appears, \ie, the target actor is placed into a new context performing the source actor's movements.
To face this problem, we propose a shape-aware human retargeting approach and a new dataset with many annotations that provide a practical benchmark to assess the progress of both retargeting and body reenactment methods.

Our dataset comprises different videos with annotated human-to-object interactions and paired motions to evaluate the task of synthesizing videos of people. This dataset can be used as a common base to improve tracking the progress of the field by showing where current approaches might fail and measuring the quality of future proposals. All data were collected in order to test methods of retargeting motion in some common real conditions, more general,  challenging scenarios, and not designed according to a specific training regime. 
Our shape-aware human retargeting approach from monocular videos of people has several advantages. It might be applied in general scenes and conditions, it is compatible with existing rendering pipelines, and it is not actor specific, conversely to most end-to-end learning techniques. The retargeting method has four main components: motion estimation of the source actor, extraction of shape and appearance of the target actor, motion retargeting with spatio-temporal constraints, and image-based rendering and composition of the video. By imposing spatial and temporal constraints on the characters' joints, our method preserves features of the motion, such as feet touching the floor and hands touching a particular object in the transferring.

We performed experiments using our new dataset containing several types of motions and actors with different body shapes and heights. Our results show that an image-based rendering technique can still exhibit a competitive quality compared to the recent deep learning techniques in generic transferring tests. Our approach achieved better results compared with end-to-end learning methodologies such as the works of~\cite{wang2018vid2vid} and~\cite{Esser_2018_CVPR} in most scenarios for appearance metrics as structural similarity (SSIM), learned perceptual similarity (LPIPS), mean squared error (MSE), and Fr\'echet Video Distance (FVD). The experimental results extend the observation of~\cite{wang2020cvpr}, where they indicate that CNN-generated images are still surprisingly easy to spot. Our results suggest that retargeting strategies based on image-to-image learning still perform poorly to several motion types and transferring contexts.

Although remarkable advances have been made in neural human rendering approaches \citep{wang2018vid2vid,chan2018dance,Aberman_2018,sun2020human}, simultaneously transferring the appearance and motion of virtual characters is still a challenging problem. Most existing algorithms are restricted to perform self-transfer (or reenactment), suffering from poor generalization to changes in scale, camera viewpoint, and intrinsic parameters. Moreover, the lack of suitable datasets for performing and evaluating retargeting approaches on a common ground hampers to track the progress of the field. These limitations are the central motivation of our proposed approach. Therefore, the contribution of this paper is three-fold:

\begin{enumerate}
	\item We present a novel video retargeting technique for human motion and appearance transferring. This technique is designed to leverage invariant information along the video transferring, such as the body shape of the person, while also taking into account physical constraints of the motion in 3D and the image domain;
	
	\item We present a new benchmark dataset composed of $88$ videos, being  $32$ paired videos with annotated human-to-object interactions and paired actor motions, eight videos where each actor rotates vertically while holding an A-pose, eight four-minute videos with the subject performing random moves, and eight $15$-seconds videos with actors dancing in pairs. The set participating actors comprises eight human subjects, carefully chosen with varying body shapes, clothing styles, heights, and gender. The designed dataset and evaluation protocol are suitable for both self-transfer and cross-transfer scenarios;
	
	\item The dataset and presented results also propose to open discussions on common evaluation metrics for video retargeting schemes. In this context, our results suggest an image-based rendering technique can still exhibit a competitive quality when compared against recent end-to-end deep learning techniques in challenging and more general video-to-video human retargeting contexts.
\end{enumerate}

\section{Related Work}\label{sec:rel}

\noindent\textbf{Human appearance transfer and image-to-image translation.} The past five years have witnessed the explosion of generative adversarial networks (GANs) to new view synthesis. The synthesis is formulated as being an end-to-end learning problem~\citep{CNN_for_view_synthesis_0,CNN_for_view_synthesis_1,CNN_for_view_synthesis_2,Balakrishnan,Esser_2018_CVPR}, where a distribution is estimated to sample new views. A representative approach is the work of~\cite{ma2017pose}, where the authors proposed to transfer the appearance of a person to a given body pose. Similarly, \cite{Lassner_GeneratingPeople} proposed ClothNet to generate images of people with similar poses and shapes in different clothing styles. \cite{Esser_2018_CVPR} used a conditional U-Net to synthesize new images based on estimated edges and body joint locations.

\begin{figure}[t!]
	\centering
	\includegraphics[width=1\linewidth]{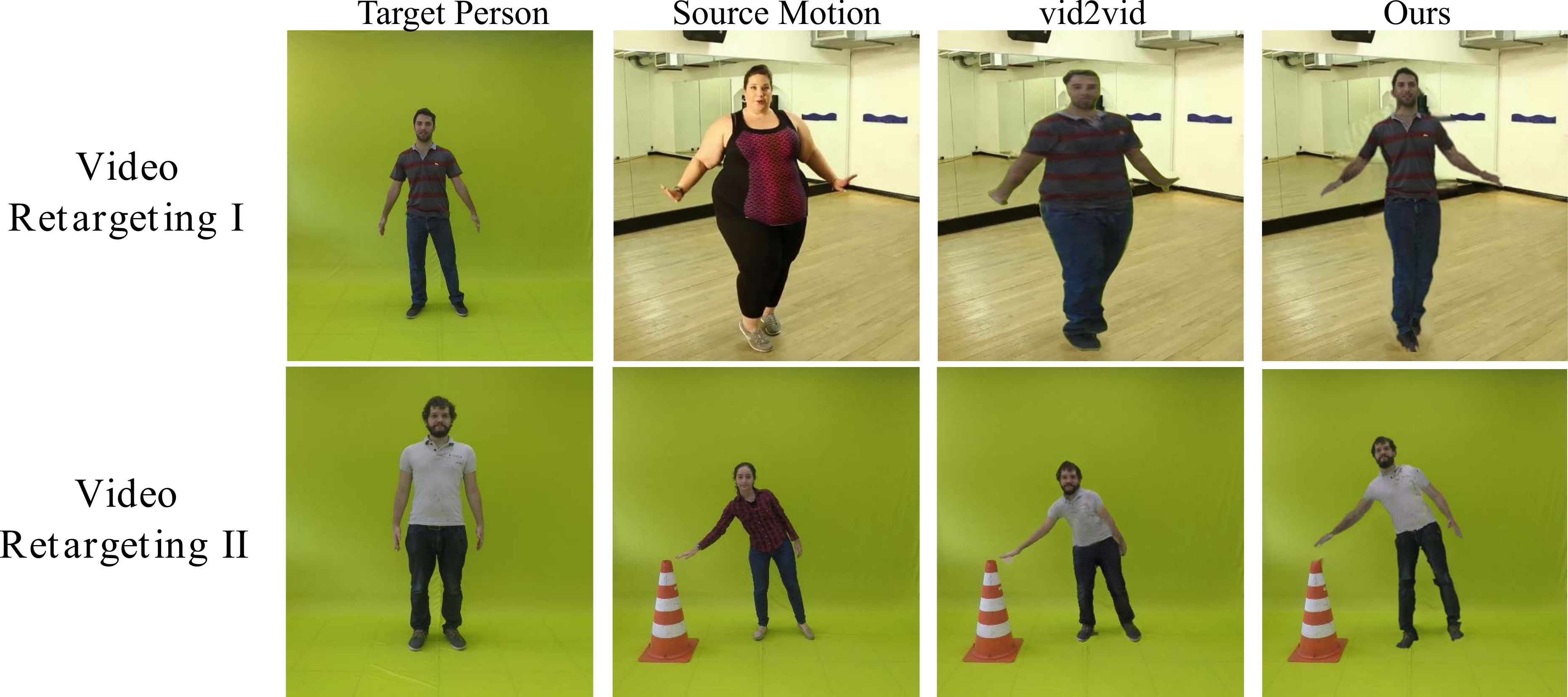}         
	\caption{\small {\bf Motion and appearance transfer in different morphologies}. From left to right: target person, source motion video with a human of different body shape, vid2vid~\citep{wang2018vid2vid}, and our retargeting results. Note that vid2vid stretched, squeezed, and shranked the body forms whenever the transferring characters have different morphologies.}
	\label{fig:ret_vid2vid}
	
\end{figure}

Recent works such as~\cite{Aberman_2018} and~\cite{chan2018dance} start applying adversarial training to map 2D poses to the appearance of a target subject. Although these works employ a scale-and-translate step to handle the difference in the limb proportions between the source skeleton and the target, their synthesized views still have clear gaps in the test time compared with the training time. \cite{wang2018vid2vid} proposed a general video-to-video synthesis framework based on conditional GANs to generate high-resolution and temporally consistent videos of people. \cite{Shysheya_2019_CVPR} attempt to handle the poor generalization by training a model using different actors' point-of-views. Their approach is also actor-specific and requires training a model for each new character, including the full acquisition setup information and camera poses. \cite{mir20} proposed to leverage the information from DensePose to learn a model to perform texture transfer of garments. Although their method is texture agnostic and not actor specific, it is designed to deal with garments transference (shirts and pants) and does not address the problem of full-body transference neither handle the cross-transference. It also disregards the motion constraints and human-to-object interactions~\citep{Hassan2019}. In the same line,~\cite{NeverovaGK18} investigated a combination of surface-based pose estimation and deep generative models; however, their method only considers the layout locations and ignores the personalized shape and limb (joint) rotations.
Despite the impressive results for several inputs, end-to-end learning-based techniques still fail to synthesize the human body's details, such as face and hands. Furthermore, it is worth noting that these techniques focus on transferring style, which leads to undesired distortions when the characters have different morphologies (proportions or body parts' lengths). An alluring example is depicted in Figure~\ref{fig:ret_vid2vid}, where we perform the transfer between actors with differences in body shape (first row) and height (second row). 

Another limitation of recent approaches such as~\cite{Aberman_2018},~\cite{chan2018dance}, and~\cite{wang2018vid2vid} is that they are dataset specific, \ie, they require training a different GAN for each video of the target person with different motions to perform the transferring. This training is computationally intensive and takes several days on a single GPU. In order to overcome these limitations,~\cite{lwb2019} proposed a 3D body mesh recovery module to disentangle the pose and shape; however, their performance significantly decreases when the source image comes from a different domain from their dataset, indicating that they are also affected by poor generalization to camera viewing changes.
Recently,~\cite{sun2020human} also proposed a  dataset-specific method where projections of the reconstructed 3D human model are used to condition the GAN training, in order to maintain the structural integrity of the transfer to different poses. Nevertheless, all analyses were made in a strict setup where the person is standing parallel to the image plane, and the considered motions have reduced lateral translations.

In this paper, we point out that there is still a performance gap of recent end-to-end deep learning techniques against an image-based model when this comprises carefully designed steps for human shape and pose estimation and retargeting. These results extend the observation of the works of~\cite{bau2019seeing} and~\cite{wang2020cvpr}, where the authors observed that GANs still present limited generation capacity. While~\citeauthor{bau2019seeing}~showed that generative network models could ignore classes that are too hard at the same time producing outputs of high average visual quality,~\citeauthor{wang2020cvpr}~demonstrated that CNN-generated images are yet surprisingly easy to spot.    

\vspace*{0.3cm}
\noindent\textbf{3D human shape and pose estimation.} Significant advances have been recently developed to estimate both the human skeleton and 3D body shape from images. \cite{Sigal} estimate shape by fitting a generative model, the SCAPE~\citep{Anguelov_2005}, to image silhouettes.~\cite{Bogo_2016} proposed the SMPLify method, which is a fully automated approach for estimating 3D body shape and pose from 2D joints in images. SMPLify uses a CNN to estimate 2D joint locations and then it fits an SMPL body model~\citep{Loper_2015} to these joints. \cite{Lassner_2017} used the curated results from SMPLify to train $91$ keypoint detectors. Similarly,~\cite{kanazawaHMR18} used unpaired 2D keypoint annotations and 3D scans to train an end-to-end network to infer the 3D mesh parameters and the camera pose.~\cite{kolotouros2019spin} combined an optimization method and a deep network to design a method less sensitive to the optimization initialization. Even though their method outperformed the works of \citeauthor{Bogo_2016}, \citeauthor{Lassner_2017}, and \citeauthor{kanazawaHMR18} regarding 3D joint error and runtime, their bounding box cropping strategy does not allow motion reconstruction from poses, since it frees three-dimensional pose regression from having to localize the person with scale and translation in image space. Moreover, they lack global information and temporal consistency in shape, pose, and human-to-object interactions, which are required in video retargeting with consistent motion transferring.

\vspace*{0.3cm}
\noindent\textbf{Retargeting motion.} \cite{Gleicher}'s seminal work of retargeting motion addressed the problem of transferring motion from one virtual actor to another with different morphologies. \cite{motion_retargetting_1} pushed further \citeauthor{Gleicher}'s method by presenting a real-time motion retargeting approach based on inverse rate control. Both~\citeauthor{Gleicher}'s and~\citeauthor{motion_retargetting_1}'s approaches require an iterative optimization with hand-designed activation constraints trajectories over time for several particular motions (like jumping or walking). Moreover, these constraints are sometimes hard to be designed if the 3D information of the environment is not explicitly available, such as in monocular videos. 

\cite{Villegas_2018_CVPR} proposed a kinematic neural network with an adversarial cycle consistency to remove the manual step of detecting the motion constraints. In the same direction, the recent work of~\cite{2018-TOG-SFV} takes a step towards automatically transferring motion between humans and virtual humanoids. Similarly,~\cite{retargeting_2d} proposed a 2D motion retargeting using a high-level latent motion representation. Their method has the benefit of not explicitly reconstructing 3D poses and camera parameters, but it fails to transfer motions if the character walks towards the camera or with variations of the camera's point-of-view. Differently from~\citeauthor{Gleicher},~\citeauthor{motion_retargetting_1},~\citeauthor{Villegas_2018_CVPR},~and \citeauthor{2018-TOG-SFV}, we propose a retargeting formulation that does not assume that all the desired constraints should be inferred only from 3D end-effectors pose, but also from the 2D locations in the image where the body-environment interactions happen (please see Figure~\ref{fig:constraits_2D}). Furthermore, our spatio-temporal modeling allows a competitive runtime during the retargeting, conversely to \citeauthor{Gleicher}'s work whose optimization can last several hours, even for a small batch of images.

Thus, inspired by the promising ideas employed to adapt motion from one character to another, 3D shape and pose estimation, our proposed methodology brings forth the advantages of jointly modeling 3D motion, shape, and appearance information. However, differently from the works mentioned above, our approach does not require training a model for each actor, keeps visual details from the target actor, does not require strict setups, and preserves the features of the transferred motion as well as the human-object interactions in the video.

\vspace*{0.3cm}
\noindent\textbf{Existing datasets.}
Datasets are one cornerstone of recent advances in Computer Vision. While there are many large datasets available for human shape and pose estimation~\citep{MPII_dataset,COCO_dataset,AMASS:ICCV:2019}, existing motion retargeting datasets are yet rare, hampering progress on this area.~\cite{Villegas_2018_CVPR} provided human joint poses from synthetic paired motions, however, paired visual information is not available, limiting the appearance transferring.~\cite{chan2018dance} made available videos with random actor movements that can be used to learn the appearance of the target actor. However, the provided data is limited to their setup requirements, and it does not allow the analysis and comparison with other methods.~\cite{lwb2019} presented a set of videos with random actions of target subjects, as well as videos of the subjects performing an A-pose movement. This set enables methods focusing on modeling 3D human body estimation or using few keyframes to be executed using their data. On the other hand, the lack of paired motions limits motion and appearance retargeting results in quantitative terms, where the source and target actors are different subjects. Conversely, our proposed dataset, in addition to videos with random actions of the target subjects and videos of the subjects performing an A-pose video, also provides several carefully paired reconstructed 3D motions and annotated human-to-object interactions in the image and 3D space.

\section{Retargeting Approach}

\begin{figure*}[t!]
	\includegraphics[width=1.0\linewidth]{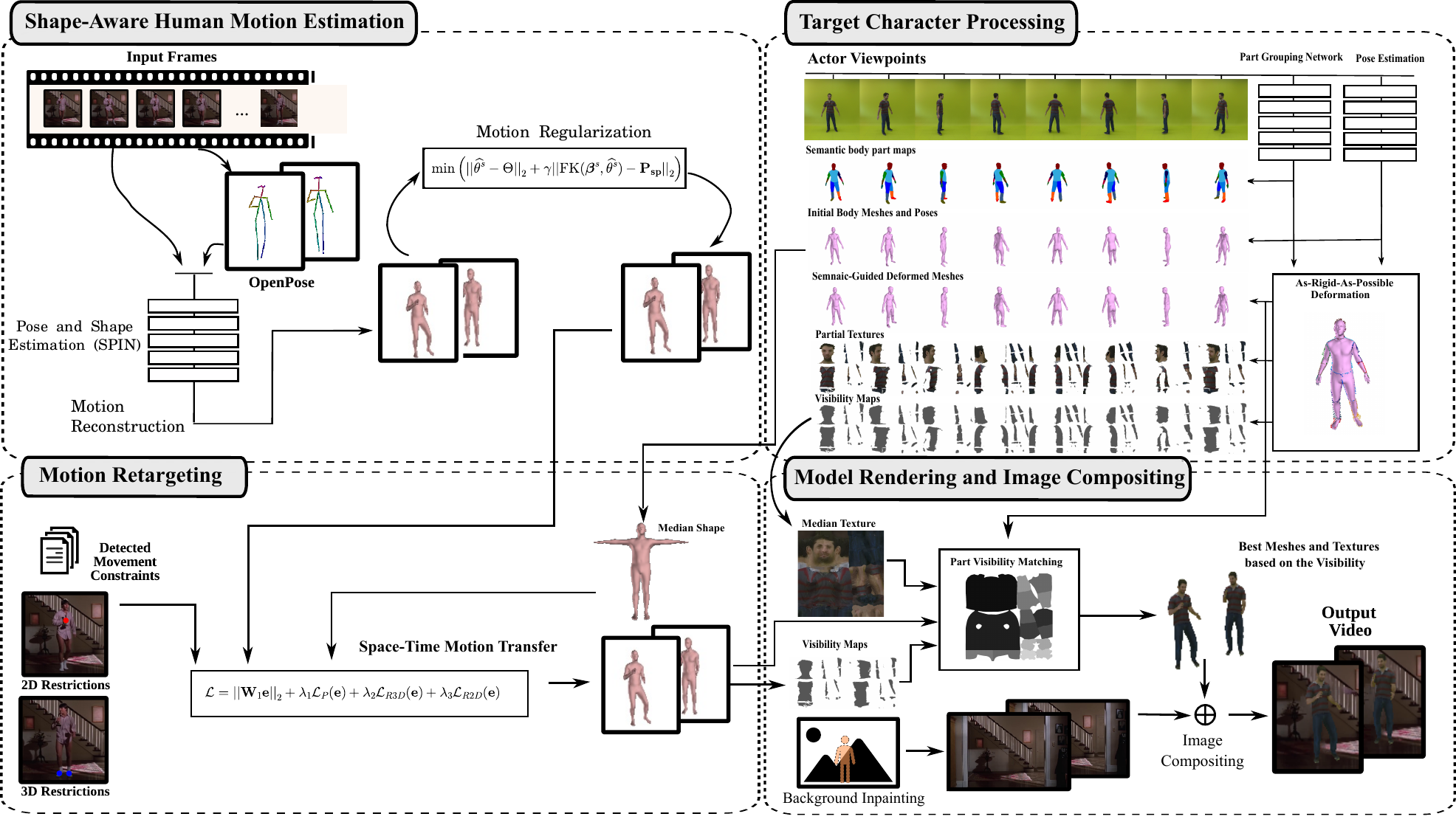}
	\caption{\textbf{Overview of our retargeting approach}. Our method is composed of four main components: human motion estimation in the source video (first component); we retarget this motion into a different target character (second component), considering the motion constraints (third component), and by last, we synthesize the appearance of the target character into the source video.}
	
	\label{fig:method}
\end{figure*}

This section presents our human transferring method considering the importance of human motion, shape, and appearance in the retargeting. Unlike most techniques that transfer either appearance~\citep{Esser_2018_CVPR,Aberman_2018,chan2018dance,wang2018vid2vid} or motion independently~\citep{Villegas_2018_CVPR,2018-TOG-SFV}, we present a new method that simultaneously considers body shape, motion retargeting constraints, and human-to-object interactions over time, while retaining visual appearance quality.

Our method comprises four main components. We first estimate the motion of the character performing actions in the source video, where essential aspects of plausible movements, such as a shared coordinate system for all image frames and temporal motion smoothness are ensured. Second, we extract the~\text{body shape and texture} of the target character in the second video. We also extract the texture, refine body geometry and estimate the visibility information of the body parts for transferring the appearance. Then, the~\textit{retargeting} component adapts the estimated movement to the body shape of the target character while considering temporal motion consistency and the physical human interactions (constraints) with the environment. Finally, the~\textit{image-based rendering and composition} component combines classical geometry rendering and image-based rendering to render the texture (appearance extracted from the target character) into the background of the source video. Figure~\ref{fig:method} shows a schematic representation of the method. 

\vspace*{0.3cm}
\noindent\textbf{Human body and motion representation.}
We represent the human motion by a set of translations and rotations over time of the joints that specify a human skeleton. This skeleton is attached to the actor's body and is defined as a $24$ linked joints hierarchy. The $i$-th joint pose $\mbf P^i \in \mathbb{SE}(3)$ is given by recursively rotating the joints of the skeleton tree, starting from the root joint and ending in its leaf joints, \ie, the forward kinematics denoted as FK. To represent the 3D shape of the human body, we adopted the SMPL model parametrization~\citep{Loper_2015}, which is composed of a learned human shape distribution $\mathcal{M}$, 3D joint angles ($\boldsymbol{\theta} \in \mathbb{R}^{72}$ defining 3D rotations of the skeleton joint tree), and shape coefficients $\boldsymbol{\beta} \in \mathbb{R}^{10}$ that model the proportions and dimensions of the human body.

\subsection{Shape-Aware Human Motion Estimation}\label{method:motion_estimation}

We start estimating the actor's motion in the source video. Our method builds upon the SPIN, a learning-based human model estimation framework of \cite{kolotouros2019spin}, where the human pose and body shape are predicted in the coordinate system of the person’s bounding box computed by OpenPose~\citep{openpose1,openpose2,openpose3}. SPIN leverages the advantages of regression and optimization-based human pose estimation frameworks to provide an efficient and accurate SMPL human model estimate. This bounding box normalizes the person in size and position. As also observed by~\cite{Mehta_2017}, the bounding box normalization frees 3D pose estimation from the burden of computing the scale factor (between the body shape to the camera distance) and the location in the image. However, this normalization incurs in a loss of temporal pose consistency required in the motion transfer. The loss of consistency also often leads to wrong body shape estimates for each frame, which should be constant through the video.

In order to overcome these issues, we map the initial pose estimation using virtual camera coordinates, following the motion reconstruction strategy presented by~\cite{Gomes_2020_WACV}. The loss in motion reconstruction is composed of two terms. While the first term enforces the pose projections of the joints to remain in the same locations into the common reference coordinate system, the second term favors maintaining the joints' angles configuration reinforcing the character shape to have averaged shape coefficients ($\boldsymbol{\beta^s}$) in the entire video. The pose of the human model in each frame is then obtained with the forward kinematics (FK) from the obtained joints configuration $\mth_k^s$ of our shape-ware motion reconstruction:
\begin{equation}
\left[\mbf P^0_{k} ~ \mbf P^1_{k} ~ \ldots~ \mbf P^{23}_{k}\right] = \mbox{FK}(\mbt^s,\mth_k^s),
\end{equation}
\noindent where $\mbf P_k^{i} = [\mbox{FK}(\boldsymbol{\beta}^s,\boldsymbol{\theta}_k^s)]_i$ is the pose of the $i$-th joint at frame $k$. 

The raw actor motion is defined as $\mbf M(\mbt^s,\mth^s) = [\mbf P_1 ~ \mbf P_2 ~ ... ~ \mbf P_n] \in \mathbb{R}^{24\times 4 \times 4 \times n}$, where $\boldsymbol{\beta} = [\mbt^s_1, \mbt^s_2, \ldots,\mbt^s_n] \in \mathbb{R}^{10\times n}\mbox{ and } \mth = [\mth_1^s, \mth_2^s, \ldots, \mth_n^s]  \in \mathbb{R}^{72\times n}$ are composed of the stacked $\mbt^s,\mth_k^s$ over time. 

\subsubsection{Motion Regularization} 

Since we estimate the character poses frame-by-frame, the resulting motion might present shaking motion with high-frequency artifacts in some short sections of the video. To alleviate these effects, we perform a regularization to seek a new set of joint angles $\widehat{\mth^s}$ that creates a smoother motion. After applying a cubic-spline interpolation~\citep{splines} over the joints' motion $\mbf{M}(\mbt^s,\mth^s)$, we remove the outlier joints from the interpolated spline. The final motion estimate is obtained by minimizing the cost:
\begin{equation}\label{eq:inversek}
\min\Big(||\widehat{\mth^s} - \Theta||_2 + \gamma||\mbox{FK}(\boldsymbol{\beta}^s,\widehat{\mth^s}) - \mbf P_{sp}||_2 \Big),
\end{equation}
\noindent where $\Theta$ is the subset of inlier joints, $\mbox{FK}$ is the forward kinematics, $\boldsymbol{\beta}^s$ defines the proportions and dimensions of the human body in the source video, $\mbf P_{sp}$ is the spline interpolated joint positions, and $\gamma$ is the scaling factor between the original joint angles and the interpolated positions. This strategy removes high-frequency artifacts of the joints' motion while retaining the movement features.

\subsection{Target Character Processing}\label{method:shape_appearance}

Most methods based on Generative Adversarial Networks, such as~\cite{wang2018vid2vid},~\cite{chan2018dance},~\cite{Aberman_2018}, and~\cite{sun2020human}, have emerged as effective approaches for human appearance synthesis. However, these methods still suffer in creating fine texture details, notably in some body parts as the face and hands. Besides, it is well known that these methods suffer from quality instability when applied in contexts slightly different from the original ones, \ie, a small difference in camera position, uncommon motions, pose translation, {\it etc}. These limitations motivate the proposal of our semantic-guided appearance retargeting method, which is designed to leverage visibility map information and semantic body parts to refine the initial target mesh model while keeping finer texture details in the transferring.

Thus, in order to create a more stable method and overcome the lack of details, we design a new semantic-guided image-based rendering approach that copies local patterns from input images to the correct position in the generated images. Our idea stems from using semantic information of the body (\eg, face, arms, torso locations, {\it etc.}) in the geometric rendering to encode patch positions and image-based rendering to copy pixels from the target images, and therefore maintaining texture details. This strategy estimates a generic target body model $\boldsymbol{\beta}^t$, comprising body geometry, texture, and visibility information at each frame that will be transferred to the source video in the retargeting step.

\subsubsection{Semantic-Guided Human Model Extraction}\label{method:geometry}    

When extracting the appearance and geometry of the human body, the self-occlusion of body parts and the deformable nature of the human body (and of clothes) bring challenging conditions. In order to tackle these difficulties, we propose a semantic-guided image-based rendering of body parts that explores the global and local information of the human body into the body model estimation. 

While we gathered the global geometric information from the pose and shape as discussed in Section~\ref{method:motion_estimation}, the local geometric information is extracted for each viewpoint and aligned with the contours of their semantic body parts in the image. To perform this alignment, we partitioned and computed the correspondence of the 3D body model into fourteen meaningful body part labels (face, left arm, {\it etc.}). The 2D semantic body labels are computed using the~\cite{Gong2018InstancelevelHP}'s body parsing model, which we fine-tuned using the people-snapshot dataset~\citep{alldieck2018video}. 

\begin{figure}[t!]
	\includegraphics[width=1\linewidth]{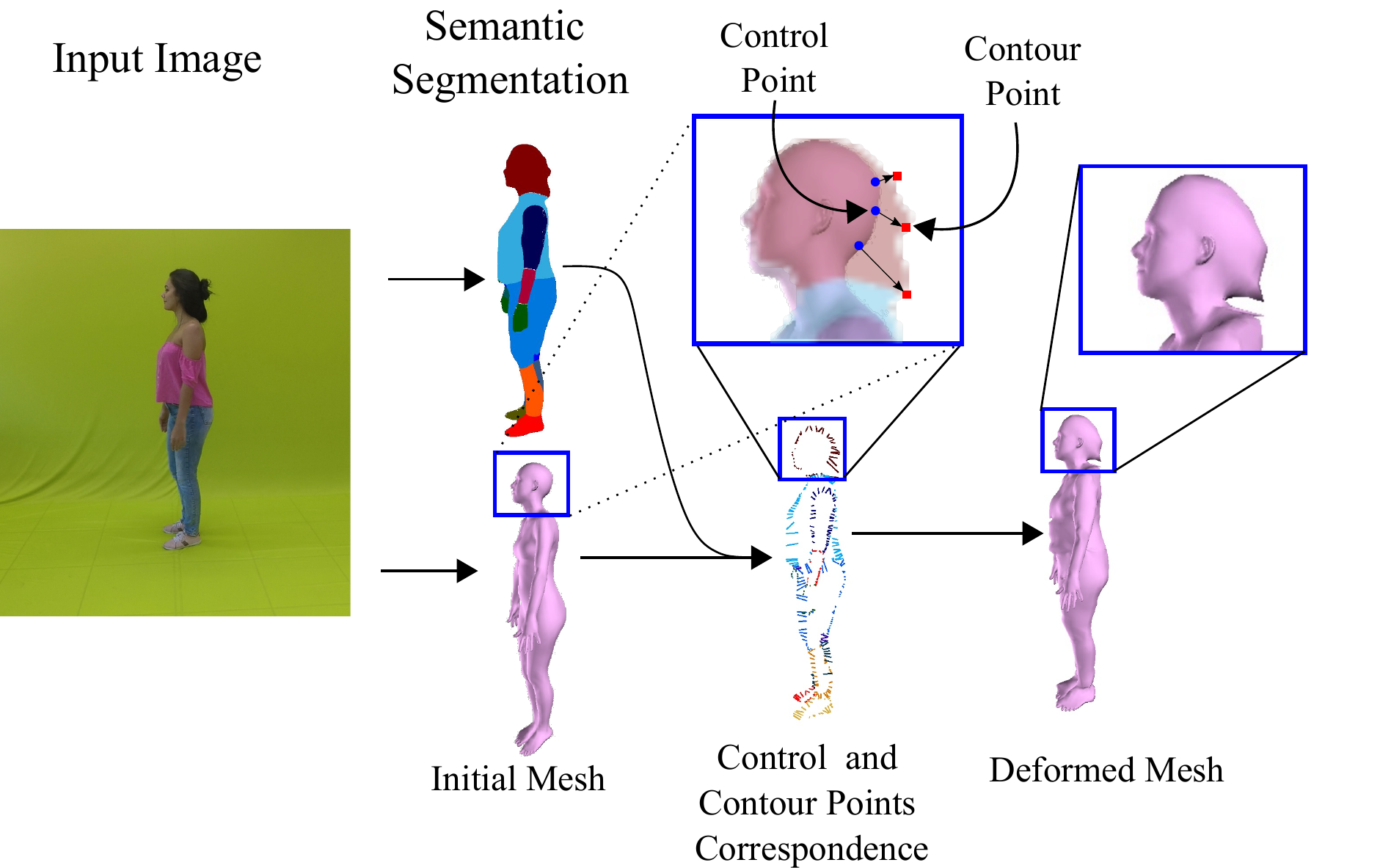}
	\caption{\textbf{Semantic guided deformation}. The contour points (red squares) in the semantic map indicate the target localization of the control points (blue circles) in the body mesh that it will guide the ARAP algorithm to fit the body mesh into the contours of the semantic map. The correspondence between each contour point and control point is illustrated with colored small lines.}
	\label{fig:deformation}
	
\end{figure}

After computing the semantic map of the body, for each contour point (red squares in Figure~\ref{fig:deformation}) in the map, we define the vertex from the body mesh with the same semantic and the smallest Euclidean distance to the contour point as a control point (blue circles in Figure~\ref{fig:deformation}). The Euclidean distance is computed between the contour point and the 2D projection of the vertex. Each control point will receive a target position given its correspondent contour point. These control points and their new locations guide the deformation of the mesh to fit the shape into the semantic map's contour. In the deformation, we seek a new body mesh $Q$ that is a locally rigid transformation of the source body mesh $P$, following the control points given by the semantic contours. The mesh deformation is solved efficiently with the local rigid deformation As-Rigid-As-Possible (ARAP)~\citep{Zohar2015}. The correspondence between each contour point and control point is represented in Figure~\ref{fig:deformation} with colored small lines. Notice that the desired motion of the control points guides the deformation to fit the body mesh into the contours of the semantic map.

\subsubsection{Human Textures and Visibility Maps Extraction}\label{method:texture} 

The geometric information allows rendering the human target character in a new viewpoint by applying the desired transformations and re-projecting them onto the image plane. In order to compare and merge the information from the human actor from different viewpoints, we map the views to a common UV texture map space. The mapping function is given by a parametric function ${\mS}$ that maps a point in the mesh from frame $k$ to a point in the texture space with coordinates $(u, v)$, $\mS:\mathbb{R}^3 \rightarrow \mathbb{R}^2$. Then, the accumulated texture map $\mUV(u,v)$ for all available images of the target character is done by the rendering with ${\mS}$:
\begin{equation}\label{eq:texture_projection}
\mUV({\mS}(x,y,z)) = \mI({\Pi}((x,y,z),\mbf K)), 
\end{equation}

\noindent where the ${\Pi}(.,\mbf K)$ operator performs the rendering taking a 3D mesh point $(x,y,z)$ and projecting it into the image plane given the camera parameters $\mbf K$, and $\mI$ is the texture information in the image coordinates. 

\begin{figure}[t]
	\includegraphics[width=1\linewidth]{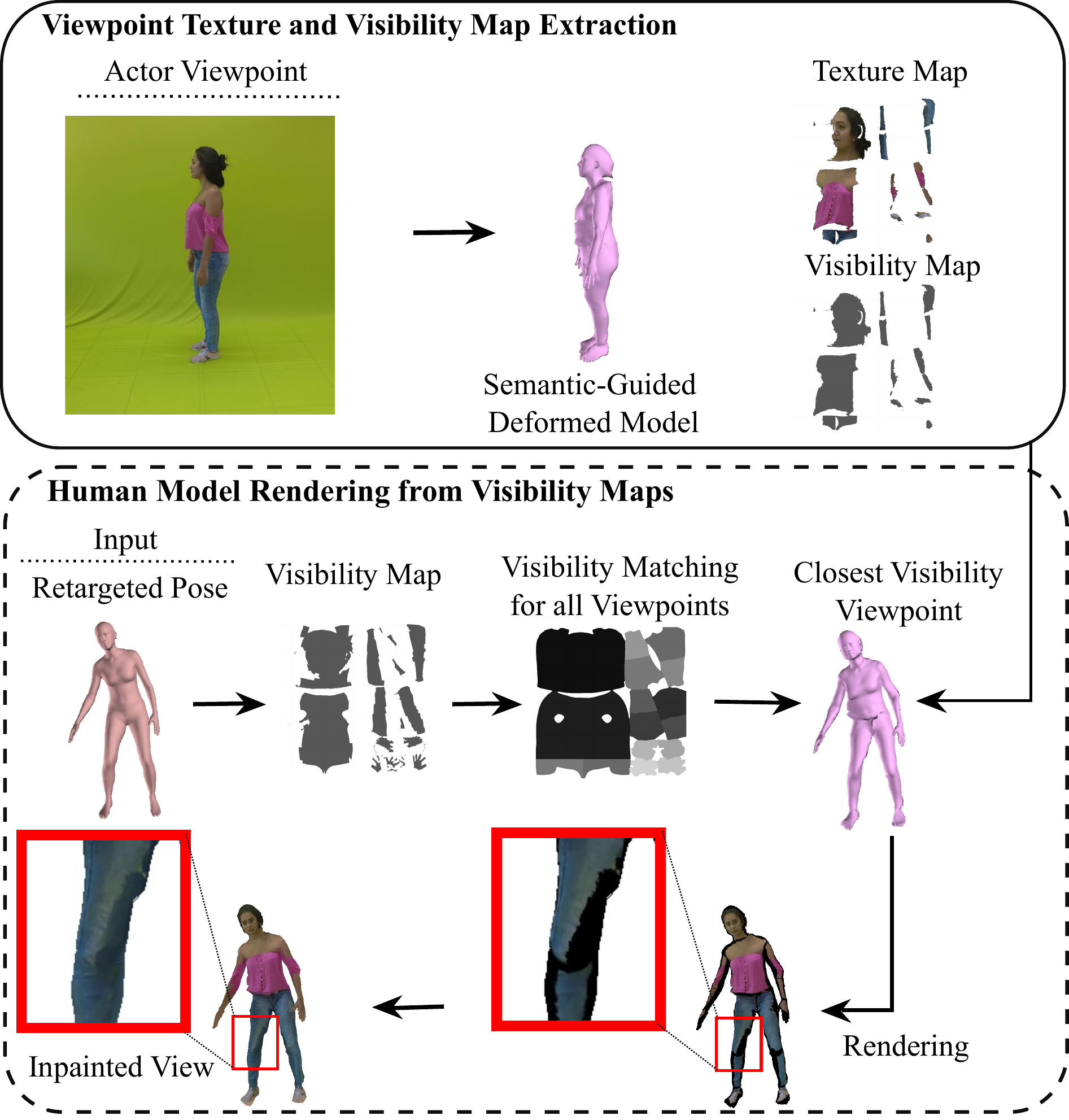}
	\caption{\textbf{Rendering of the visibility maps and texture images.} {\it Top:} We project each target actor viewpoint in a common UV texture space using the estimated geometry and create a binary map of visibility body parts. {\it Bottom:} Given the goal pose (retargeted pose), we estimate its visibility body parts map, and then select the better matching visibility body parts created from the viewpoints from the target actor.}
	\label{fig:texture-vis-map}
	
\end{figure}

Finally, we assert which mesh points are visible by exploring the inverse map ${\mS}^{-1}(u,v)$, as illustrated in Figure~\ref{fig:texture-vis-map}. Each visibility map indicates which parts of the body model are visible per frame. Then we select the closest viewpoint to the desired new viewpoint, for each part of the body model from the visibility maps.

\subsection{Motion Retargeting using Hybrid 2D/3D Spatio-Temporal Constraints}

After estimating the motion from the input video, \ie, $\mbf M$($\boldsymbol{\beta}^s,\boldsymbol{\theta}^s$), and 3D model $\boldsymbol{\beta}^t$ of the target human, we can proceed to the motion retargeting step. We assume that the target character has a homeomorphic skeleton structure to the source character, \ie, the geometric differences are in terms of bone lengths and body proportions. Our retargeting motion estimation loss is designed to guarantee the motion similarity and physical human-object interaction constraints over time. Similar to~\cite{Gleicher}, our first goal is to retain the joint configuration of the target as close as possible to the source joint configurations at instant $k$, $\mth_k^t \approx \mth_k^s$, \ie, to keep $\mbf e_k$ small such as: $\mth_k^t = \mth_k^s + \mbf e_k.$

We also aim to keep similar movement style and speed in the retargeted motion. Thus, we propose a one step speed prediction in 3D space defined as $\Delta \mbf M(\mbt,\mth_k) = \mbox{FK}(\mbt,\mth_{k+1}) - \mbox{FK}(\mbt,\mth_k)$ to maintain the motion style from the original joints' motion:

\begin{equation}\label{eq:pred}
\loss_P(\mbf e) = \sum_{k=i+1}^{i+n}||\Delta \mbf M(\mbt^t, \mth_{k}^s + \mbf e_k) - \Delta \mbf M(\mbt^s, \mth_{k}^s) ||_1,
\end{equation}
\noindent where $\mbf e = [\mbf e_{i+1}, \ldots, \mbf e_{i+n}]^T$, and $n$ is the number of frames considered in the retargeting. 

Rather than considering a loss for the total number of frames, we use only the frames belonging to a neighboring temporal window of $n$ frames equivalent to two seconds of video. This neighboring temporal window scheme allows us to track the local temporal motion style producing a motion that tends to be natural compared with a realistic-looking of the estimated source motion. The retargeting considering a local neighboring window of frames also results in a more efficient optimization.

\begin{figure}[t!]
	\includegraphics[width=1\linewidth]{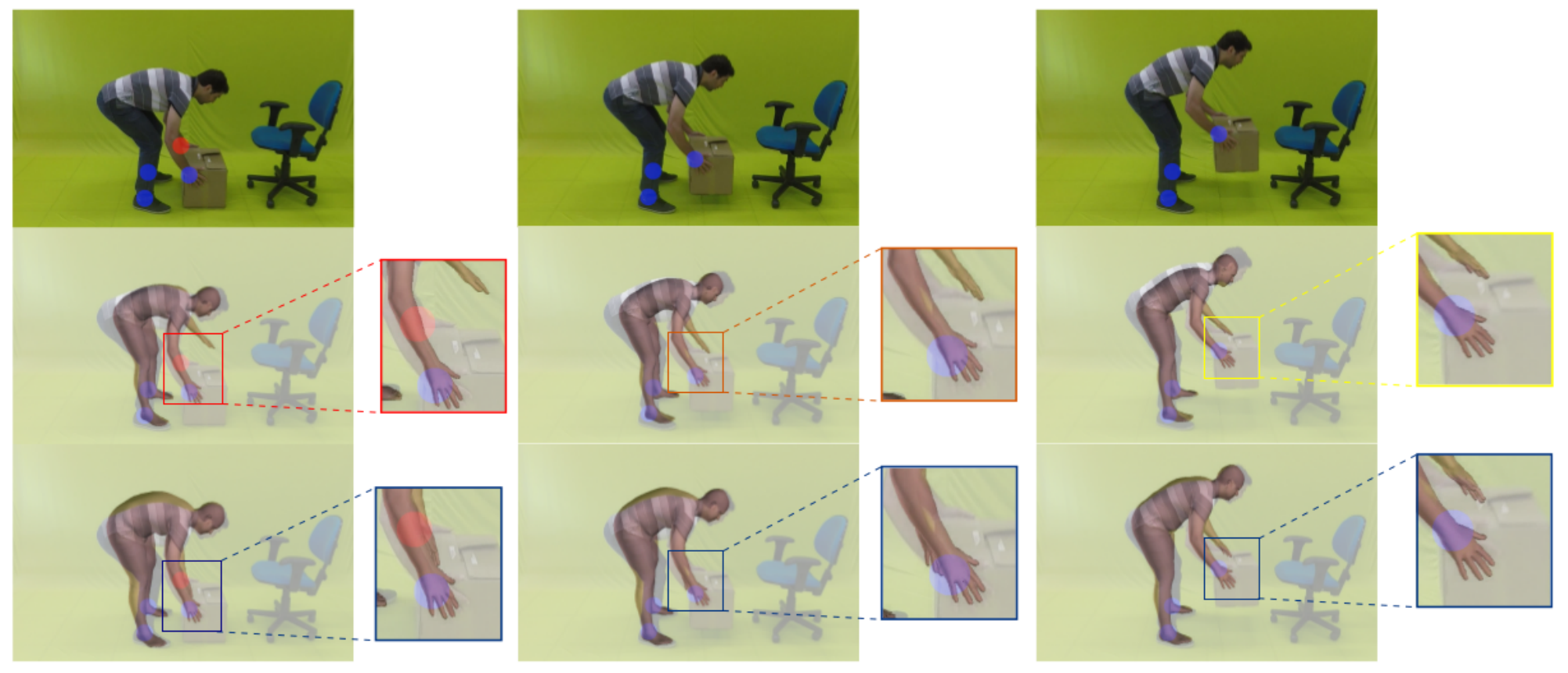}
	\caption{\textbf{Example of hybrid 2D/3D constraints from human-to-object interactions.} {\it Top row:} Original video with 3D constraints (blue dots) and 2D constraints (red dots). {\it Middle row:} Motion retargeting to a new character using only 3D constraints. {\it Bottom row:} The results for the new character applying hybrid 2D/3D constraints using our retargeting approach. Please observe that the hands' positions are more consistent when adopting our hybrid strategy.}
	\label{fig:constraits_2D}
\end{figure}

\vspace*{0.3cm}
\noindent\textbf{2D/3D human-to-object interactions.} 
The human-to-object interactions (\ie, motion constraints) are important to identify key features of the original motion that must be preserved in the retargeted motion. The specification of these interactions typically involves only a small amount of work in comparison with the task of creating new motions. Typical interactions are, for instance, that the target character's feet should be on the floor; holding hands while dancing or while grabbing/manipulating an object in the source video. Some examples of human-to-object motion constraints are shown in Figure~\ref{fig:constraits_2D}, where the actor is interacting with a box. 

Going one step further than classic retargeting constraints defined in~\cite{Gleicher} and \cite{motion_retargetting_1}, where end-effectors must be at solely a desired 3D position at a given moment, we propose an extended hybrid constraint in the image domain, \ie, the joint of the character must also be projected at a specific location in the image. This type of motion constraint allows the user to exploit the visual knowledge of interactions of the actor in the scene. Some examples are shown in Figure~\ref{fig:constraits_2D}, where two types of constraints are defined: 3D interactions (blue dots) impose the feet and right hand to be in the same location after the retargeting, and 2D constraints (red dots) imposing the correct position to the left hand in the image.    

Our retargeting is capable of adapting to such situations by defining the motion retargeting constraints losses in respect to end-effectors' (hands, feet) 3D poses $\mbf P_{R3D}$ and 2D poses $\mbf P_{R2D}$ as:

\begin{align}
&\loss_{R3D}(\mbf e_k) = ||\mbox{FK}(\mbt^t,\mth_{k}^s + \mbf e_k) - \mbf P_{R3D}||_1,\label{eq:constr} \\
&\loss_{R2D}(\mbf e_k) = ||\Pi(\mbox{FK}(\mbt^t,\mth_{k}^s + \mbf e_k), \mbf K) - \mbf P_{R2D}||_1. \label{eq:constr_2d}
\end{align}

\vspace*{0.3cm}
\noindent\textbf{Space-time loss optimization.} The final motion retargeting loss $\loss$ combines the source motion appearance with the different shape and constraints of the target character from Equations~\ref{eq:pred}, \ref{eq:constr}, and \ref{eq:constr_2d}: 
\begin{equation}
\loss = \mbf ||\mbf W_1\mbf e||_2 + \lambda_1\loss_P(\mbf e) + \lambda_2\loss_{R3D}(\mbf e) + \lambda_3 \loss_{R2D}(\mbf e),
\end{equation}
\noindent where the joint parameters to be optimized are $\mbf e = [\mbf e_{i+1}, \ldots, \mbf e_{i+n}]^T$, $n$ is the number of frames considered in the retargeting window, $\lambda_1$, $\lambda_2$, and $\lambda_3$ are the contributions for the different error terms, and $\mbf W_1$ is a positive diagonal matrix of weights for the motion appearance for each body joint. This weight matrix is set to penalize more errors in joints that are closer to the root joint. 

One representative example of the retargeting strategy considering these hybrid 2D/3D constraints is shown in Figure~\ref{fig:constraits_2D}. In this video sequence, the target actor is bigger and taller than the one in the source video (shown in the two first rows of the figure). Notice that the retargeting of the target actor (shown in the third row) results in more bent poses to maintain the human-to-object interactions, and thus the hands' positions are consistent when adopting our strategy.

\subsection{Model Rendering and Image Compositing}\label{method:render}

In our framework's last step, we combine the rendered target character and the source background to make a convincing final image. We first segment the source image into a background layer using, as a mask, the projection of the actor body model with a dilation, and then the background is inpainted. 

Our experiments showed that different inpainting techniques, including end-to-end learning-based methods such as~\cite{yu2018free},~\cite{yu2018generative},~\cite{wang2018videoinp}, and~\cite{Xu_2019_CVPR}, displayed small differences in the quality of results, probably because most part of the filled region will be covered by the target person. Thus, we apply the method proposed by~\cite{based-inpainting} that presented the best overall results, especially when the background texture is homogeneous. To ensure temporal smoothness in the inpainting background, we compute the pixel color value as the median value between the neighboring $n$ frames. 

In the sequence, we explored the visibility map of the retargeted body model (global geometric information discussed in Section~\ref{method:texture}) to select the human body parts that better matches the target parts. Since the transformation between the retargeted SMPL model and the estimated SMPL model to the images is known, we apply the same transformation used in the local geometric information to move them to the correct positions. Instead of directly applying 3D warping in the selected images, we use our pre-warping step (texture map) to improve the rendering speed of 3D warping. In order to fill the remaining part holes in the warped image, we explored the median of the accumulated texture $\mUV$ (see Figure~\ref{fig:texture-vis-map}). Finally, the background and the target character are combined in the retargeted frame.

\section{Human Retargeting Dataset and Evaluation}

Due to the lack of suitable benchmark datasets, recent works on neural view synthesis and body reenacting~\citep{chan2018dance,Esser_2018_CVPR,wang2018vid2vid} only analyze their results in qualitative terms or quantitative terms to self-transfer in which the source and target actors are the same subjects. The provided data is adapted to the requirements of their method setup, and it is hard to perform a comparison with other methods on this data. Cross-transfer is far more complicated than self-subject transfer. First, self-transfer is not affected by body appearance changes (from body shape, clothing). Second, the self-shape transfer does not account for human-to-object interactions or disregards the influence of existing human-environment physical interactions. Moreover, as previously discussed in Section~\ref{sec:rel}, existing video retargeting datasets are still rare.

\subsection{Human Retargeting Dataset}
\begin{figure*}[t!]
	\centering
	\includegraphics[width=1.0\linewidth]{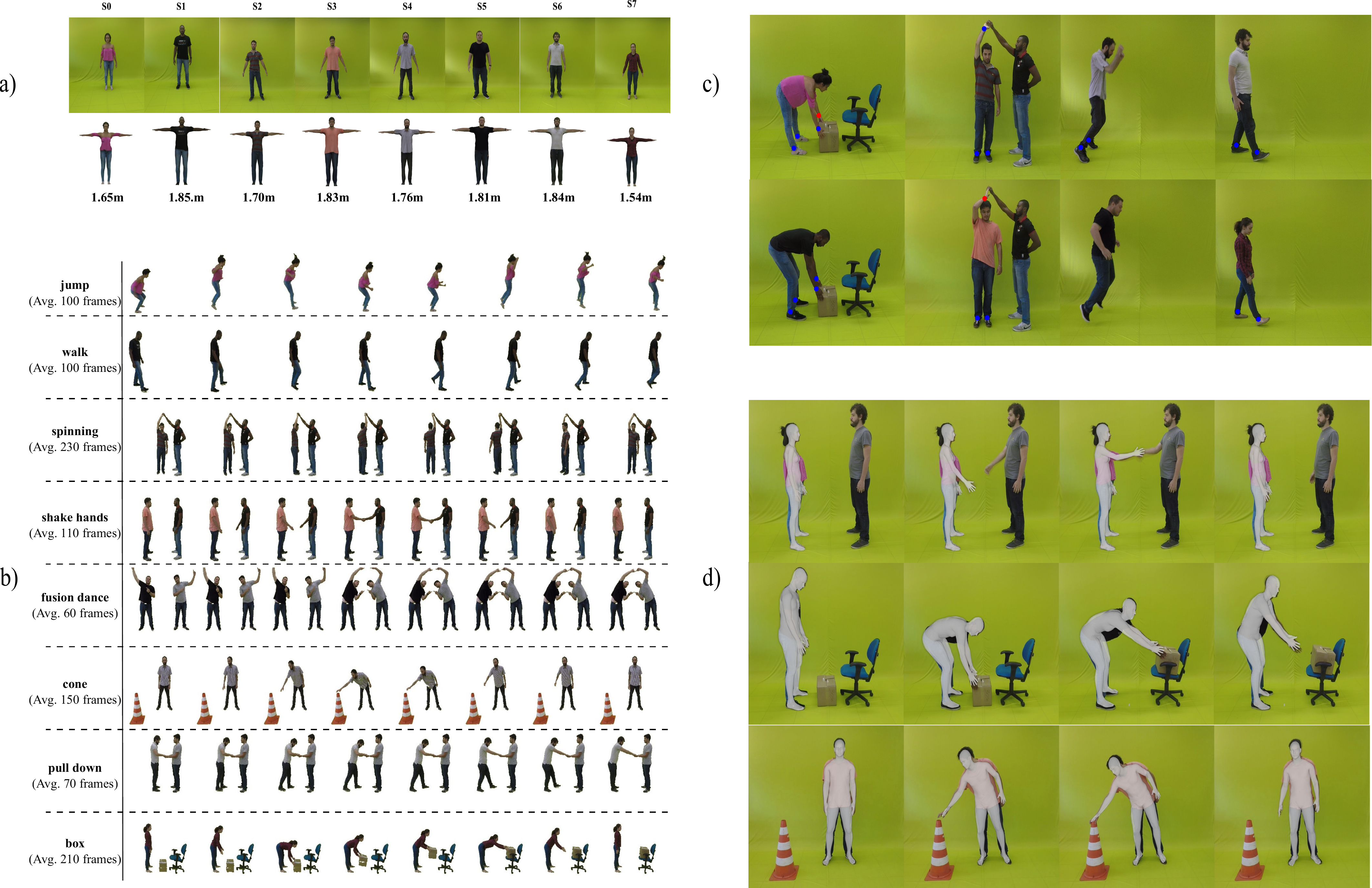}         
	\caption{\textbf{Human retargeting dataset.} {\it a)} The subjects participating in our dataset, their respective height and estimated SMPL body models. {\it b)} Overview of all motions available in our proposed dataset. {\it c)} Paired motions (upper and lower rows) with annotated motion constraints (3D constraints in blue and 2D constraints in red). {\it d)} The reconstructed 3D motions.}    
	\label{fig:dataset}
	
\end{figure*}  

To evaluate the retargeting and appearance transfer with different actor motions, consistent reconstructed 3D motions, and with human-to-object interactions, we created a new dataset with {\it paired motion} sequences from different characters and {\it annotated motion retargeting constraints}. For each video sequence, we provide a refined 3D actor reconstructed motion and the actor body shape estimated with~\cite{alldieck2018video}. The refined reconstructed 3D motions and 2D-3D annotation of interactions were collected by manual annotation. The provided motions are not prone to typical motion artifacts such as bended knees, body shape variations, and camera-to-actor translation changes.

Our carefully designed dataset comprises $8$ subjects. To keep the dataset with diversity, we choose participants (subjects S0 to S7) with different gender, sizes, clothing styles, and body shapes. Figure~\ref{fig:dataset} shows the subjects, their respective body models, and labels. We also choose a set of movements that are representative of the problem of the retargeting, increasing the level of difficulty and with different motion constraints. These movements are: ``pick up a box'', ``spinning'', ``jump'', ``walk'', ``shake hands'', ``touch a cone'', ``pull down'', and ``fusion dance''.

Each actor performed all eight actions. We paired two actors to perform the same motion sequence, where the subjects were instructed to follow marks on the floor to perform the same action, resulting in four paired videos per action (a total of $32$ paired videos). Then, we define the combination of actors aiming at the most challenging configuration for the task of human retargeting motion. For instance, actors $S0$, $S2$, $S4$, and $S6$ were paired respectively with $S1$, $S3$, $S5$, and $S7$. We also provide, for each subject, three videos: one video where the subject is rotating and holding an A-pose, a four-minute video where the subject is performing different poses, and a $15$-second video where the subject is dancing. All videos were recorded with $1{,}000 \times 1{,}080$ of resolution at $30FPS$. This information allows training most existing approaches for evaluation. Figure~\ref{fig:dataset} shows some examples of frames from our dataset.

Additionally, we provide three publicly available videos acquired under uncontrolled conditions: {\it joao-pedro}, {\it tom-cruise}, and {\it bruno-mars} sequences. The {\it bruno-mars} sequence was also adopted in the experiments of~\cite{chan2018dance}. The provided videos present more general conditions to analyze the effectiveness of reenacting and retargeting methods in the wild with more complex motions, background, and varying illumination.

\subsection{Protocol}

The evaluation often employed by works on character reenacting/synthesis~\citep{Aberman_2018,chan2018dance,lwb2019,sun2020human} consists in setting the source character equal to the target character. However, we argue that this protocol is used because of the absence of paired motion sequences, as also noted in~\cite{lwb2019} and~\cite{sun2020human}. While this protocol might be appropriate to assess new synthesized view/pose in the same scene background, we state that it is not appropriate for evaluating the synthesis of new videos of people when taking into account motion constraints (\eg, human-to-object interactions), distinct shapes and heights, and transferring them in different backgrounds where they were initially recorded. 

Therefore, we run our evaluation protocol as follows: when evaluating a new synthesized video, we place the target actor performing a similar motion to the source actor (all physical constraints and the human-to-object interactions are taken into account). Then, we move the target actor to a different scenario configuration to perform the retargeting. If the retargeting is successfully executed, the method will place the target actor moving as the source actor into the source actor's scenario.

\subsection{Evaluation Metrics}

We measure the quality of the synthesized frames in terms of the following metrics: i) the structural similarity (SSIM)~\citep{Wang04imagequality} that compares local patterns of pixel intensities normalized for luminance and contrast. SSIM assumes that human visual perception is highly adapted for extracting structural information from objects; ii) learned perceptual similarity (LPIPS)~\citep{zhang2018perceptual}, which provides a deep neural learned similarity distance closer to human visual perception; iii) the mean squared error (MSE) that is computed by averaging the squared intensity differences between the pixels; and iv) Fr\'echet video distance (FVD)~\citep{unterthiner2019accurate}, which was designed to capture the quality of the synthesized frames and their temporal coherence in videos. These are widely used perceptual distances to measure how similar the synthesized images are in a way that coincides with human judgment. 

To properly evaluate the quality of retargeting, it is required a paired video where the target character is the same as the source video sequence. Collecting real paired motion sequences from different characters is a challenging task, even when the movement of the actors is carefully predefined and synchronized. For instance, each actor has a movement style that can result in videos with unsynchronized actions (as seen in the third column in Figure~\ref{fig:dataset}-c). Thus, to make possible the computation of quantitative metrics, we relax the assumption that the frame $k$ in the synthesized video must be the frame $k$ in the paired video by applying a small window of acceptance around $k$, \ie, we evaluate the quality of one synthesized frame as a better answer between the frame and a small window of frames in the paired video. The window of acceptance $w$ is estimated for each pair of videos according to: 
\begin{equation}\label{eq:window_size}
w(V_1, V_2)  = \max(15, 2\times( \abs{len(V_1) - len(V_2)} ),
\end{equation}

\noindent where $len(V)$ is the number of frames in the video $V$. Equation~\ref{eq:window_size} captures how much two videos are not synchronized allowing the synthesized frames to match with the paired video. The lower bound value of $15$ frames was empirically selected by a visual analysis of our videos.

Aside from the image quality metrics, we also propose to evaluate the approaches with fake image detectors. Since our dataset provides paired motion sequences, we can evaluate the retargeted frames' quality and realism with an image forgery detector. We applied the detector in the generated and real frames from the sequence where the same subject performs the retargeted motion. The retargeted frames were generated using motion extracted from source videos whose subject performing the motion is different from the target subject. 

We adopted the image forgery detection algorithm presented by~\cite{marra2019e2e} to evaluate all methods in our experiments. The~\citeauthor{marra2019e2e}'s method evaluates images by detecting pixel artifacts, which is a common metric for detecting CNN-generated images. We remark that we tested different image forgery detection algorithms such as the method proposed by~\cite{Wang2019CNNgeneratedIA}, but the results were inconclusive, which might be because of the high resolution of our images. Furthermore, our images are not only generated by CNN methods, and as the authors stated, their method performs at chance in visual fakes produced by image-based rendering or classical commercial rendering engines. Finally, we define the forgery performance as the difference between the probability of the paired real image and its respective synthetic frame of being classified as fake. This difference indicates how far a synthetic frame is from being recognized as fake with a similar result of a real image.

\section{Experiments and Results}

\noindent\textbf{Parameter settings.} We selected a fixed set of parameters for all experiments. We applied a grid search over a portion of the $4$-minutes videos training examples of the dataset. The grid search $\gamma$ values ranged from $5$ to $100$ with a step of $5$, and $\lambda$ from $1$ to $10$, with a step of $1$. In the human motion estimation and retargeting steps, we used $\gamma = 10$, $\lambda_1 = 5$, $\lambda_2 = 1$, and $\lambda_3 = 1$. We minimize the retargeting loss function with Adam optimizer using $300$ iterations, learning rate $0.01$, $\beta_1=0.9$, and $\beta_2=0.99$. To deform the target body model with the semantic contour control points, we employed the default parameters proposed by \cite{Zohar2015}, and we fed the method with eight images taken from different viewpoints of the actor (see Figure~\ref{fig:method}). For a complete comparison, we also present qualitative results of the new frames from the state-of-art approaches by replacing the generated background regions from the methods by the source video background.

\begin{table}[t!]
	\centering
	\caption{{\bf Ablation study}. Comparison of mean MSE and FVD for different ablated versions of our method on all motion types of our dataset (best in bold).}
	\label{table:ablation_results}
	
	\resizebox{0.99\columnwidth}{!}{%
		
		\begin{tabular}{@{}lrrrr@{}}
			\toprule 
			
			{\bf Method} & \phantom{a} & {\centering MSE$\downarrow$}
			&  {\centering FVD$\downarrow$} \\ \midrule
			
			No motion regularization & \phantom{a} &$275.47$ & $887.94$ \\ 
			No semantic guidance & \phantom{a} &$275.88$ & $879.86$ \\ 
			No 2D/3D constraints & \phantom{a} &$285.33$ & $859.67$ \\ 				
			No use of visibility mask & \phantom{a} &$273.45$ & $789.06$ \\ 
			Full method & \phantom{a} &$\mathbf{259.75}$ & $\mathbf{738.96}$ \\ 	
			\bottomrule
			
		\end{tabular}
		
	}
	
\end{table}

\begin{table*}[t!]
	\centering
	\caption{{\bf Quantitative ablation analysis of the retargeting.} Table shows the error in pixels between the constraints position and the end-effectors location. This error indicates how far the end-effector is from the target position (better close to $0$).}
	\label{table:ablation_retargeting}
	
	\resizebox{0.99\linewidth}{!}{%
		
		\begin{tabular}{@{}lccccccccccccccc@{}}
			\toprule 
			\multirow{3}{*}{\bf Method} & \multicolumn{9}{c}{{\bf Motion type}} & \phantom{a} & \multicolumn{5}{c}{{\bf Pair of actors}}\\ \cmidrule{2-10} \cmidrule{12-16} 
			\multicolumn{1}{r}{} 
			&  {\centering jump}
			&  {\centering walk}
			&  {\centering spinning}
			&  {\centering shake hands}
			&  {\centering cone}
			&  {\centering fusion dance}
			&   {\centering pull down}
			&  {\centering box}
			&  {\centering Avg.} 
			& \phantom{a}
			&  {\centering S0-S1} 
			&  {\centering S2-S3} 
			&  {\centering S4-S5}  
			&  {\centering S6-S7} 
			& {\centering Avg.} \\ \midrule

			Direct Transfer & $13.73$ & $14.49$ & $21.90$ & $18.22$ & $16.77$ & $24.67$ & $34.11$ &  $22.49$ & $20.80$ & \phantom{a}
			& $27.41$ & $22.84$ & $10.06$ & $22.88$ & $20.80$  \\
			
			Retargeting & $2.44$ & $1.94$ & $2.04$ & $2.90$ & $2.76$ & $10.66$ & $3.14$ & $5.35$ & $\mathbf{3.90}$ & \phantom{a}
			& $4.57$ & $3.19$ & $2.67$ & $5.19$ & $\mathbf{3.90}$  \\

			\bottomrule \\
		\end{tabular}
		
	}

\end{table*}

\begin{figure*}[t!]
	\centering
	\includegraphics[width=0.8\linewidth]{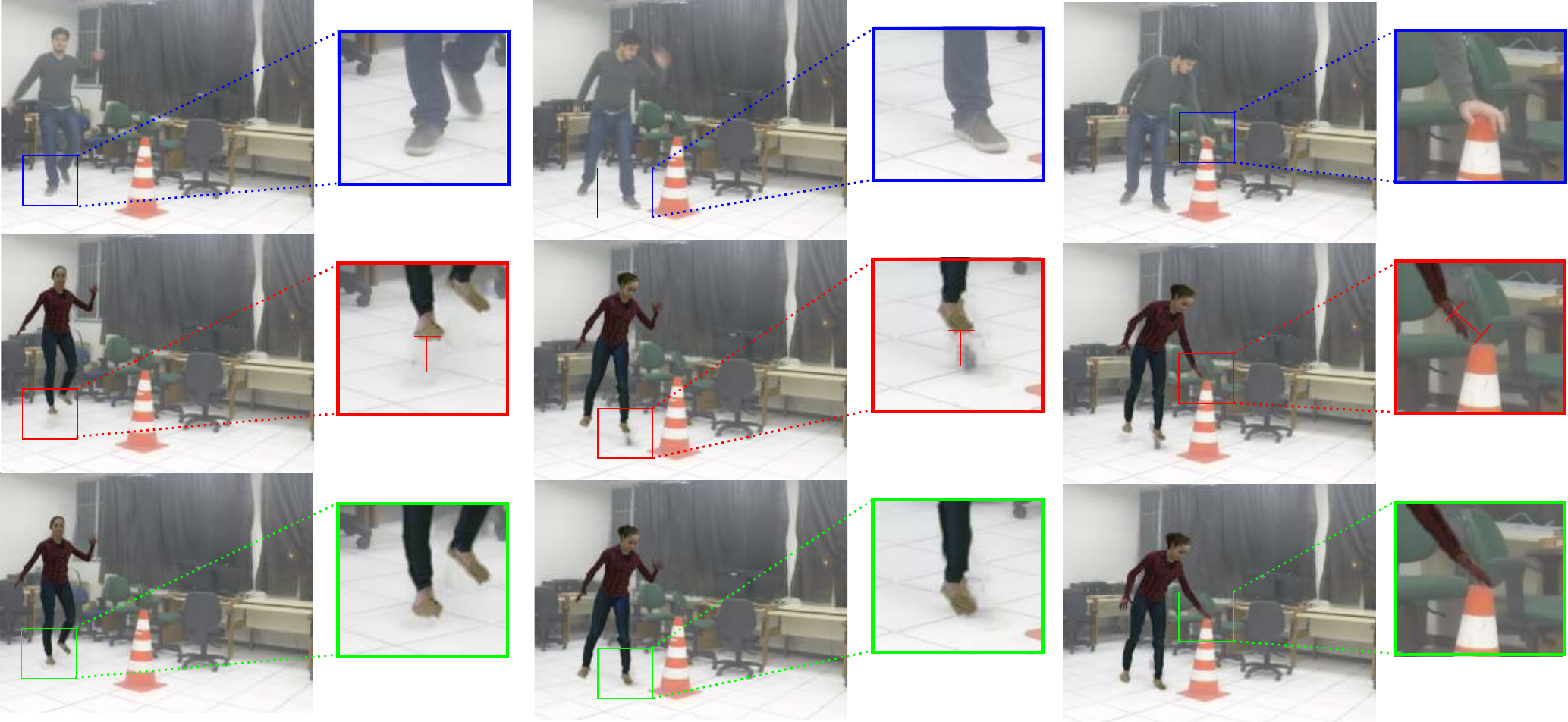}
	\caption{\textbf{Qualitative ablation analysis of the retargeting.} {\it Top row:} source video containing the actor S3. {\it Middle row:} transferring results to the actor S7 without the physical interactions. {\it Bottom row:} transferring results of our method with 2D-3D interactions.}
	\label{result:ablation-cone}
\end{figure*}

\subsection{Ablation Study}

To verify that the space-time motion transfer optimization, motion regularization, semantic-guided deformation, and visibility maps contribute to our approach's success in producing more realistic frames, we conducted several ablation analysis using the motion sequences from our dataset.  

Table~\ref{table:ablation_results} shows the results of our ablation study in terms of MSE and FVD of five ablated versions of our method. We draw the following observations. First, the best result is achieved when the full method is applied. Second, removing the 2D/3D constraints reduces the performance in terms of MSE. These constraints play a key role in the compliance of the poses to motion constraints. By removing them, large fragments from the background are computed as part of the retarget character body when computing MSE using the paired video, leading to the worst MSE value. Third, without the shape-aware regularization, which hinders the temporal coherence, the model presents the worst value of FVD. We can also see that after removing the semantic guidance, which decreases the quality of the texture applied onto the 3D model, the frames have more artifacts, and the model also performs poorly in terms of FVD.

\begin{figure}[t]
	\includegraphics[width=1\linewidth]{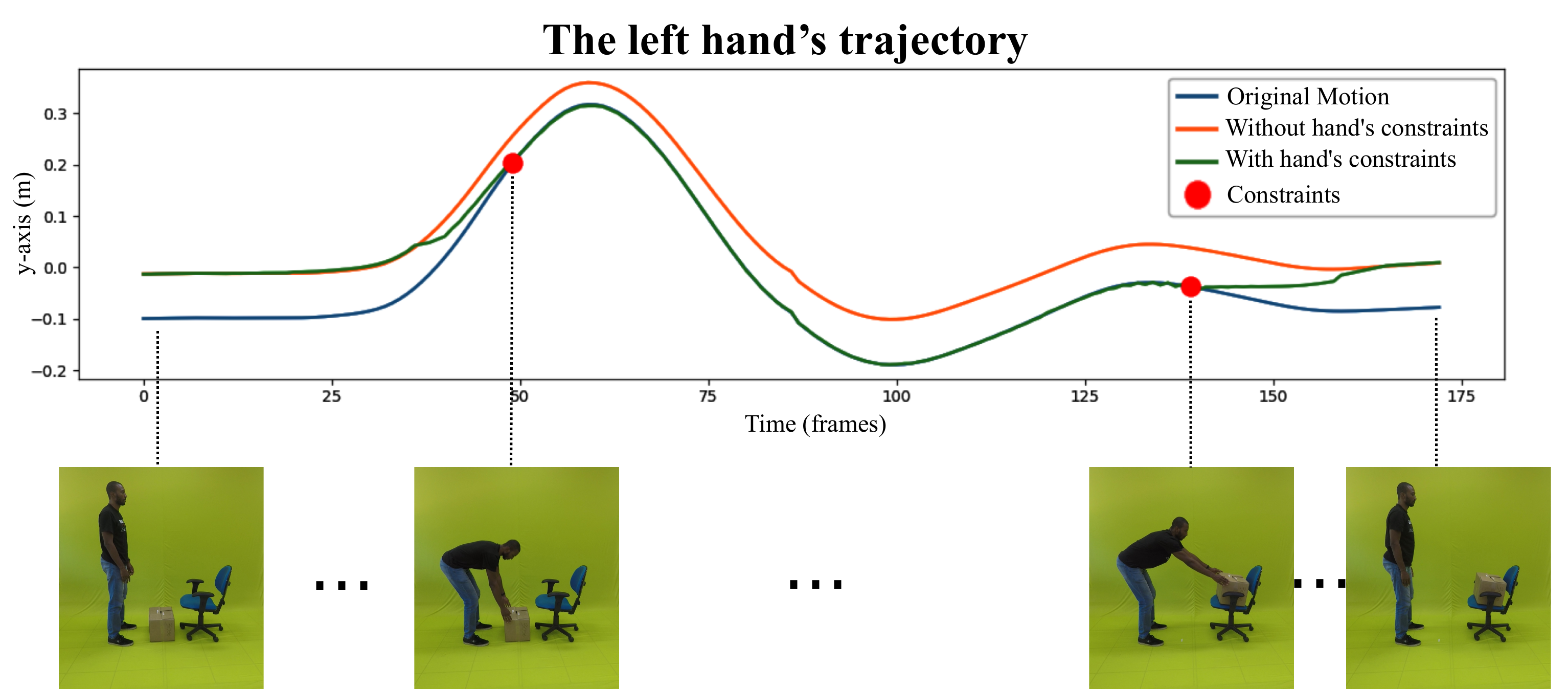}
	\caption{\textbf{Retargeted trajectory with motion constraints.} The curves show the left hand's trajectory on the y-axis when transferring the motion of {\it picking up a box} between two differently sized characters: original motion (blue line), a na\"ive transfer without constraints at the person's hand (red line), and with constraints (green line). Frames containing motion constraints are located between the red circles.}
	\label{fig:grafico_retarget}
\end{figure}

The results of a more detailed performance assessment of the effects from motion constraints in the video retargeting are shown in Table~\ref{table:ablation_retargeting}. The error between the computed position of the end-effectors and the target positions (motion constraints from the human-to-object interactions) is significantly smaller for all motion sequences and pairs of actors when applying our retargeting strategy. Some frames are shown in Figure~\ref{result:ablation-cone} to illustrate the created visual artifacts when not considering the motion constraints. In this setup, the source actor (top row) is taller than the target (bottom row), and the target actor does not touch the floor nor touch with her hands the cone without the hybrid motion constraints (middle row). Conversely, these features are kept when considering the 2D/3D human-to-object losses in the retargeting. Another representative example of the retargeted motion trajectory over time, with one shorter actor interacting with a box, is shown in Figure~\ref{fig:grafico_retarget}. Please notice the smooth motion adaptation produced by the retargeting with the restrictions in frames $47$ and $138$ (green line) when the character's left hand is touching the box. Additionally, these results illustrate that our method is able to impose different space-time human-to-object interactions and motion constraints in the retargeting.

Table~\ref{table:ablation_study} shows the forgery performance when synthesizing the frames after removing the visibility map extraction and the semantic-guided human model extraction. We can see that these two steps significantly enhance the quality of the results to a point where the detector returns for the ``shake hands" sequence a probability of $29.24$ higher when removing these two components.

\begin{table}[t!]
	\centering
	\caption{{\bf Visibility maps and semantic-guidance analysis}. Average forgery performance for each movement (better close to $0$).}
	\label{table:ablation_study}
	
	\scriptsize
	\resizebox{\columnwidth}{!}{
		\begin{tabular}{@{}lccr@{}}
			\toprule
			
			\multirow{2}[3]{*}{Motion}  & \multicolumn{2}{c}{Method} \\
			\cmidrule(l{2pt}r{2pt}){2-3} & \vtop{\hbox{\strut No Semantic-guidance}\hbox{\strut and Visibility Map}}  & \vtop{\hbox{\strut Complete}\hbox{\strut Model}} \\ \midrule
			\multicolumn{1}{l}{jump} & $46.97$ & $\mathbf{31.11}$\\
			\multicolumn{1}{l}{walk}  & $45.76$ & $\mathbf{23.35}$\\
			\multicolumn{1}{l}{spinning} & $34.46$ & $\mathbf{21.09}$\\
			\multicolumn{1}{l}{shake hands}  & $34.91$ & $\mathbf{5.67}$\\
			\multicolumn{1}{l}{cone}  & $39.20$ & $\mathbf{24.80}$\\
			\multicolumn{1}{l}{fusion dance}  & $33.24$ & $\mathbf{15.97}$\\
			\multicolumn{1}{l}{pull down}  & $35.68$ & $\mathbf{17.85}$\\
			\multicolumn{1}{l}{box} & $38.09$ & $\mathbf{17.63}$\\
			
			\bottomrule
		\end{tabular}
	}
\end{table}

\subsection{Comparison with Previous Approaches}

\begin{table*}[t!]
	\centering
	\caption{{\bf Comparison with state of the art}. SSIM, LPIPS, MSE, and FVD comparison by motion types and pair of actors from our dataset (best in bold, second-best in italic).}
	\label{table:metrics_result}
	
	\resizebox{0.99\linewidth}{!}{%
		
		\begin{tabular}{@{}clrrrrrrrrrrrrrrr@{}}
			\toprule 
			\multirow{3}{*}{\bf Metric} & \multirow{3}{*}{\bf Method} & \multicolumn{9}{c}{{\bf Motion type}} & \phantom{a} & \multicolumn{5}{c}{{\bf Pair of actors}}\\ \cmidrule{3-11} \cmidrule{13-17} 
			& \multicolumn{1}{r}{} 
			&  {\centering jump}
			&  {\centering walk}
			&  {\centering spinning}
			&  {\centering shake hands}
			&  {\centering cone}
			&  {\centering fusion dance}
			&   {\centering pull down}
			&  {\centering box}
			&  {\centering Avg.} 
			& \phantom{a}
			&  {\centering S0-S1} 
			&  {\centering S2-S3} 
			&  {\centering S4-S5}  
			&  {\centering S6-S7} 
			& {\centering Avg.} \\ \midrule
			
			\multirow{5}{*}{\rotatebox[origin=c]{90}{\parbox[c]{1.5cm}{\centering SSIM$\uparrow$}}} 
			
			& V-Unet & $0.870$ & $0.871$ & $0.843$ & $0.847$ & $0.862$ & $0.797$ & $0.847$ &  $0.857$ & $0.849$ & \phantom{a}
			& $0.855$ & $0.886$  & $0.830$ & $0.826$ & $0.849$  \\
			
			& Vid2Vid & $0.880$ & $0.884$ & $0.856$ & $0.858$ & $0.878$ & $0.821$ & $0.859$ &  $0.866$ & $\mathit{0.862}$ & \phantom{a}
			& $0.868$ & $0.901$  & $0.848$ & $0.835$ & $0.862$  \\
			
			& EBDN & $0.878$ & $0.880$ & $0.855$ & $0.859$ & $0.878$ & $0.820$ & $0.857$  & $0.858$ & $0.861$ & \phantom{a}
			& $0.867$ & $0.898$  & $0.844$ & $0.834$ & $0.861$  \\
			
			& iPER & $0.877$ & $0.880$ & $0.852$ & $0.859$ & $0.877$ & $0.816$ & $0.855$ & $0.856$ & $0.859$ & \phantom{a} 
			& $0.867$ & $0.896$  & $0.842$ & $0.831$ & $0.859$  \\
			
			& Ours & $0.881$ & $0.885$ & $0.855$ & $0.860$ & $0.879$ & $0.820$ & $0.861$ & $0.869$ & $\mathbf{0.864}$ & \phantom{a}
			& $0.872$ & $0.902$  & $0.846$ & $0.834$  & $\mathbf{0.864}$  \\
			
			\midrule
			
			\multirow{5}{*}{\rotatebox[origin=c]{90}{\parbox[c]{1.5cm}{\centering LPIPS$\downarrow$}}} 
			
			& V-Unet & $0.147$ & $0.132$ & $0.157$ & $0.161$ & $0.174$ & $0.243$ & $0.166$ &  $0.158$ & $0.167$ & \phantom{a}
			& $0.184$ & $0.160$  & $0.166$ & $0.158$ & $0.167$  \\
			
			& Vid2Vid & $0.131$ & $0.105$ & $0.126$ & $0.136$ & $0.133$ & $0.203$ & $0.142$ &  $0.133$ & $\mathit{0.138}$ & \phantom{a}
			& $0.148$ & $0.131$  & $0.129$ & $0.147$ & $0.138$  \\
			
			& EBDN & $0.141$ & $0.122$ & $0.139$ & $0.138$ & $0.143$ & $0.215$ & $0.151$  & $0.170$ & $0.153$ & \phantom{a}
			& $0.159$ & $0.145$  & $0.147$ & $0.159$ & $0.153$  \\
			
			& iPER & $0.151$ & $0.134$ & $0.151$ & $0.151$ & $0.155$ & $0.239$ & $0.168$ & $0.184$ & $0.167$ & \phantom{a} 
			& $0.161$ & $0.165$  & $0.170$ & $0.171$ & $0.167$  \\
			
			& Ours & $0.125$ & $0.099$ & $0.130$ & $0.131$ & $0.128$ & $0.206$ & $0.131$ & $0.127$ & $\mathbf{0.135}$ & \phantom{a}
			& $0.133$ & $0.129$  & $0.133$ & $0.143$  & $\mathbf{0.135}$  \\
			
			\midrule			
			
			\multirow{5}{*}{\rotatebox[origin=c]{90}{\parbox[c]{1.5cm}{\centering MSE$\downarrow$}}} 
			
			& V-Unet & $295.04$ & $269.59$ & $354.33$ & $377.02$ & $328.68$ & $559.75$ & $417.00$ &  $346.13$ & $368.44$ & \phantom{a}
			& $381.77$ & $344.71$  & $362.63$ & $384.66$ & $368.44$  \\
			
			& Vid2Vid & $257.33$ & $206.42$ & $286.18$ & $332.09$ & $253.04$ & $452.77$ & $349.28$ &  $288.54$ & $\mathit{303.32}$ & \phantom{a}
			& $312.40$ & $274.80$  & $269.81$ & $356.26$ & $\mathit{303.32}$  \\
			
			& EBDN & $306.92$ & $269.58$ & $312.69$ & $312.12$ & $266.17$ & $463.79$ & $384.23$  & $361.57$ & $334.63$ & \phantom{a}
			& $324.40$ & $314.10$  & $331.83$ & $368.19$ & $334.63$  \\
			
			& iPER & $313.43$ & $275.22$ & $344.94$ & $314.92$ & $267.39$ & $504.00$ & $404.79$ & $377.16$ & $350.23$ & \phantom{a} 
			& $277.98$ & $328.07$  & $358.30$ & $436.57$ & $350.23$  \\
			
			& Ours & $237.16$ & $178.57$ & $286.86$ & $270.25$ & $237.86$ & $434.64$ & $301.66$ & $245.86$ & $\mathbf{274.11}$ & \phantom{a}
			& $243.95$ & $260.14$  & $294.88$ & $297.46$  & $\mathbf{274.11}$  \\
			
			\midrule
			\multirow{5}{*}{\rotatebox[origin=c]{90}{\parbox[c]{1.5cm}{\centering FVD$\downarrow$}}} 
			
			& V-Unet & $1,491.63$ & $845.44$ & $1,721.81$ & $1,257.20$ & $1,415.24$ & $1,712.93$ & $2,437.98$ &  $1,816.94$ & $1,587.39$ & \phantom{a}
			& $2,239.14$ & $1,352.10$  & $1,856.78$ & $1,108.34$ & $1,639.09$  \\
			
			& Vid2Vid & $879.94$ & $266.6$ & $1,085.49$ & $396.31$ & $790.79$ & $997.42$ & $997.96$ &  $1,069.85$ & $810.48$ & \phantom{a}
			& $778.53$ & $719.80$  & $762.46$ & $574.08$ & $\mathit{708.72}$  \\
			
			& EBDN & $887.56$ & $273.00$ & $918.94$ & $423.08$ & $725.49$ & $952.22$ & $1,113.46$  & $853.26$ & $\mathit{768.37}$ & \phantom{a}
			& $791.98$ & $751.45$  & $560.27$ & $826.71$ & $732.60$  \\
			
			& iPER & $1,770.31$ & $656.07$ & $1,531.64$ & $1,266.14$ & $1,051.42$ & $1,322.72$ & $1,440.94$ & $1,719.55$ & $1,344.84$ & \phantom{a} 
			& $1,270.41$ & $1,092.82$  & $1,395.81$ & $1,214.64$ & $1,243.42$  \\
			
			& Ours & $1,119.50$ & $330.91$ & $674.99$ & $478.93$ & $767.68$ & $791.01$ & $988.35$ & $760.33$ & $\mathbf{738.96}$ & \phantom{a}
			& $715.00$ & $653.30$  & $720.49$ & $515.61$  & $\mathbf{651.10}$  \\
			
			\bottomrule
		\end{tabular}
		
	}
	
\end{table*}

We compare our method against four recent representative methods with different assumptions, including V-Unet~\citep{Esser_2018_CVPR}, vid2vid~\citep{wang2018vid2vid}, EBDN~\citep{chan2018dance}, and iPER~\citep{lwb2019}. V-Unet is a notorious representative of image-to-image translation methods using conditional variational autoencoders to generate images based only on a 2D skeleton and an image from the target actor. Similar to V-Unet, iPER is not dataset-specific, but it is a generative model trained in an adversarial manner. Vid2vid and EBDN methods, for their turn, are dataset-specific methods, \ie, they require training a GAN for several days over one video of the target subject in a large set of different poses.

\vspace*{0.3cm}
\noindent\textbf{Processing time.} Although vid2vid and EBDN required a few seconds to generate a new frame, the training step of vid2vid spent approximately $10$ days on an NVIDIA Titan XP GPU for each target subject, and to run the fine-tuning of the EBDN took approximately $4$ days to complete all stages for each subject.  On the other hand, our retargeting approach does not need to be trained. The significant parts of our method's processing time are the retargeting optimization, the deformation, and the model rendering. On an Intel Core i7-7700 CPU and NVIDIA Titan XP GPU, the average run-time for one frame of retargeting optimization was about $1.2$ seconds, including I/O. The deformation took around $12$ min on $8$ frames. The model rendering, \ie, selecting the best texture map considering the visibility map, warping the selected parts, and filling all the holes in the texture, took about $30$ seconds per frame. Thus, the total processing time $t(N)$ in seconds to run our method on a video with $N$ frames with a resolution of $1{,}920 \times 1{,}080$ is approximately $t(N) = 1.2\times N + 720 + 30 \times N$.

\begin{figure*}[t!]
	\includegraphics[width=1.0\linewidth]{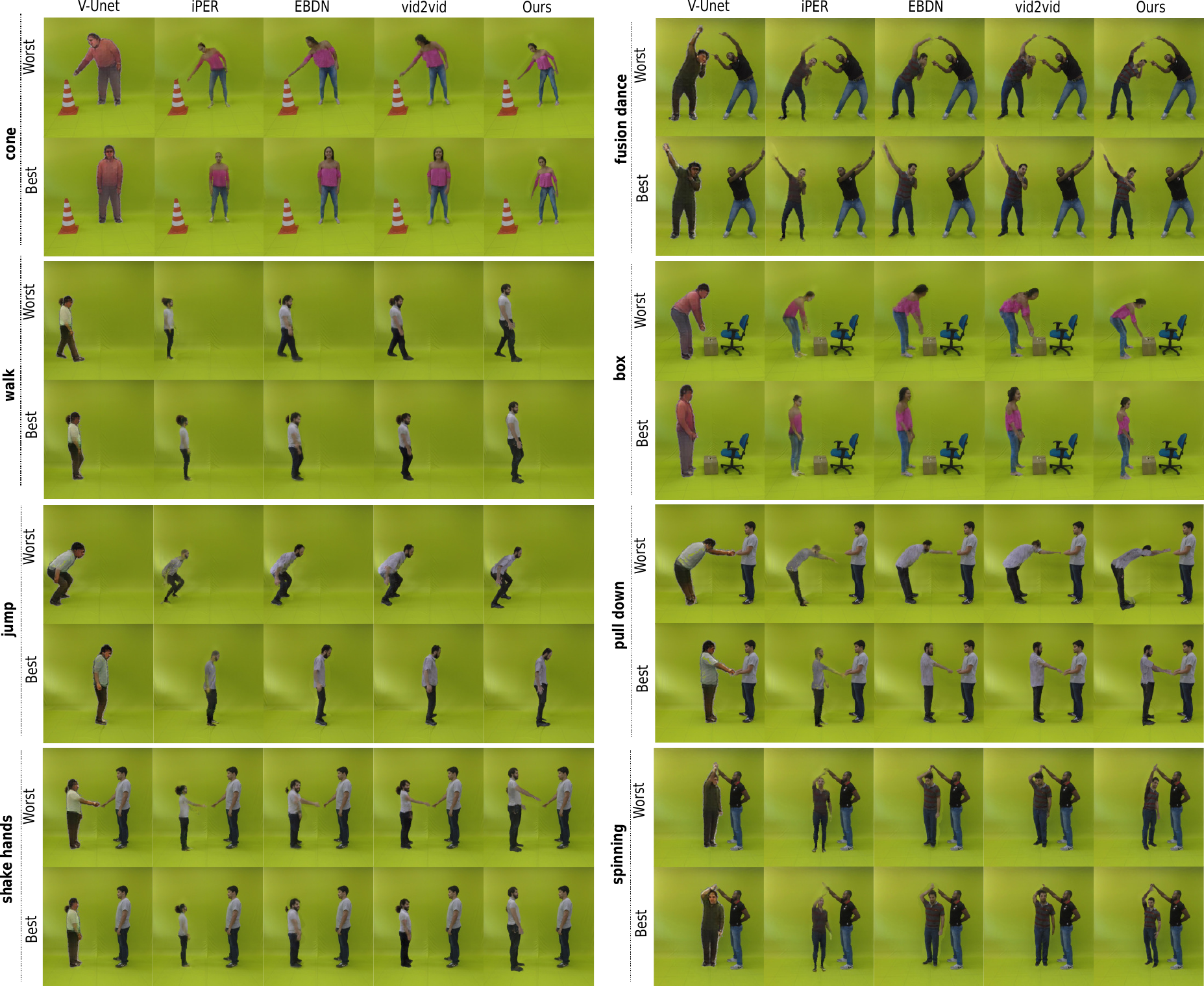}
	\caption{\textbf{Motion analysis in the dataset sequences.} Transferring results considering the cases where the person is not standing parallel to the image plane or has the arms in front of the face. In each sequence: the first row shows the worst generated frame for each method and the second row presents the best generated frame for each method.}
	\label{result:quali-dataset-0}
\end{figure*}

\subsubsection{Quantitative Analysis} 

We performed the video retargeting in all sequences in our dataset, including several public dance videos also adopted in the works~\cite{chan2018dance} and~\cite{lwb2019}. Table~\ref{table:metrics_result} shows the comparison of our approach in the dataset considering the motion types and pair of actors. We can see that despite not being dataset-specific, V-Unet and iPER did not perform well when the reference image is not from their datasets. One can see that our method outperforms, on average, all methods in considering SSIM, LPIPS, MSE, and FVD metrics. Regarding the experiments considering the pairs of actors, our approach also achieved the best average results in all metrics. In particular, our method presented better results when the subjects have different heights (S0-S1 and S6-S7). We ascribe this performance to our method being aware of the shape and physical interactions, which allows it to correct the person's position when the source actor is taller or smaller than the target person. 

We also evaluated our results (in terms of SSIM, LIPIPS, and MSE metrics) with the Wilcoxon Signed-Rank Test to check whether the difference between our results' central tendency and the baselines are statistically significant. For all baselines, except vid2vid, the test rejects the null hypothesis. In other words, the test indicates that the samples were drawn from a population with different distributions and the differences between the metrics indicate that our method performed better statistically. Moreover, it is noteworthy that the sequences ``spinning'' and ``fusion dance'' are challenging for all methods, including our methodology that was affected by wrong pose estimations. Our approach was slightly outperformed by vid2vid only in these two sequences.

\begin{table}[t!]
	\centering
	\caption{{\bf Forgery performance}. Quantitative metrics of the movement transfer approaches considering the forgery performance metric (better close to $0$).}
	
	\label{table:results_deepfake}
	\resizebox{\columnwidth}{!}{
		\begin{tabular}{@{}crrrrrrrrrrrrc@{}}
			\toprule
			
			\multirow{2}[3]{*}{Motion}  & \multicolumn{5}{c}{Method} \\
			\cmidrule(l{3pt}r{3pt}){2-6}
			& EBDN & iPER & vid2vid & V-Unet & Ours \\
			
			\midrule
			\multicolumn{0}{l}{jump} & $39.46$ & $46.56$ & $41.66$ & $47.41$ & $\mathbf{31.11}$  \\
			\multicolumn{0}{l}{walk}  & $41.00$ & $44.27$ & $30.79$ & $45.79$ & $\mathbf{23.35}$\\
			\multicolumn{0}{l}{spinning} & $33.64$ & $33.56$ & $32.36$ & $34.51$ & $\mathbf{21.09}$ \\
			\multicolumn{0}{l}{shake hands}  & $31.71$ & $33.56$ & $19.39$ & $34.93$ & $\mathbf{5.67}$ \\
			\multicolumn{0}{l}{cone}  & $38.23$ & $37.84$ & $\mathbf{24.74}$ & $39.25$ & $24.80$ \\
			\multicolumn{0}{l}{fusion dance} & $26.07$ & $32.64$ & $16.49$ & $33.38$ & $\mathbf{15.97}$ \\
			\multicolumn{0}{l}{pull down}  & $31.30$ & $34.65$ & $19.85$ & $35.67$ & $\mathbf{17.85}$ \\
			\multicolumn{0}{l}{box} & $35.96$ & $37.11$ & $24.70$ & $38.12$ & $\mathbf{17.63}$  \\
			
			\bottomrule
		\end{tabular}
	}
\end{table}

\vspace*{0.3cm}
Table~\ref{table:results_deepfake} shows the results for the experiments on image forgery detection.
These experiments indicate that the fake detector in the frames generated by our method has the closest performances to real images. For instance, in the ``shake hands'' sequence, the probability of a frame synthesized by our method to be fake is only $5.67\%$ higher than when applying the detector to the respective real frame.

\subsection{Qualitative Analysis}

We evaluated the capability of our method to transfer motion and appearance and retain interactions of the original motion despite the target actor having different proportions to the actor in the source video.

\begin{figure}[!t]
	\includegraphics[width=0.95\linewidth]{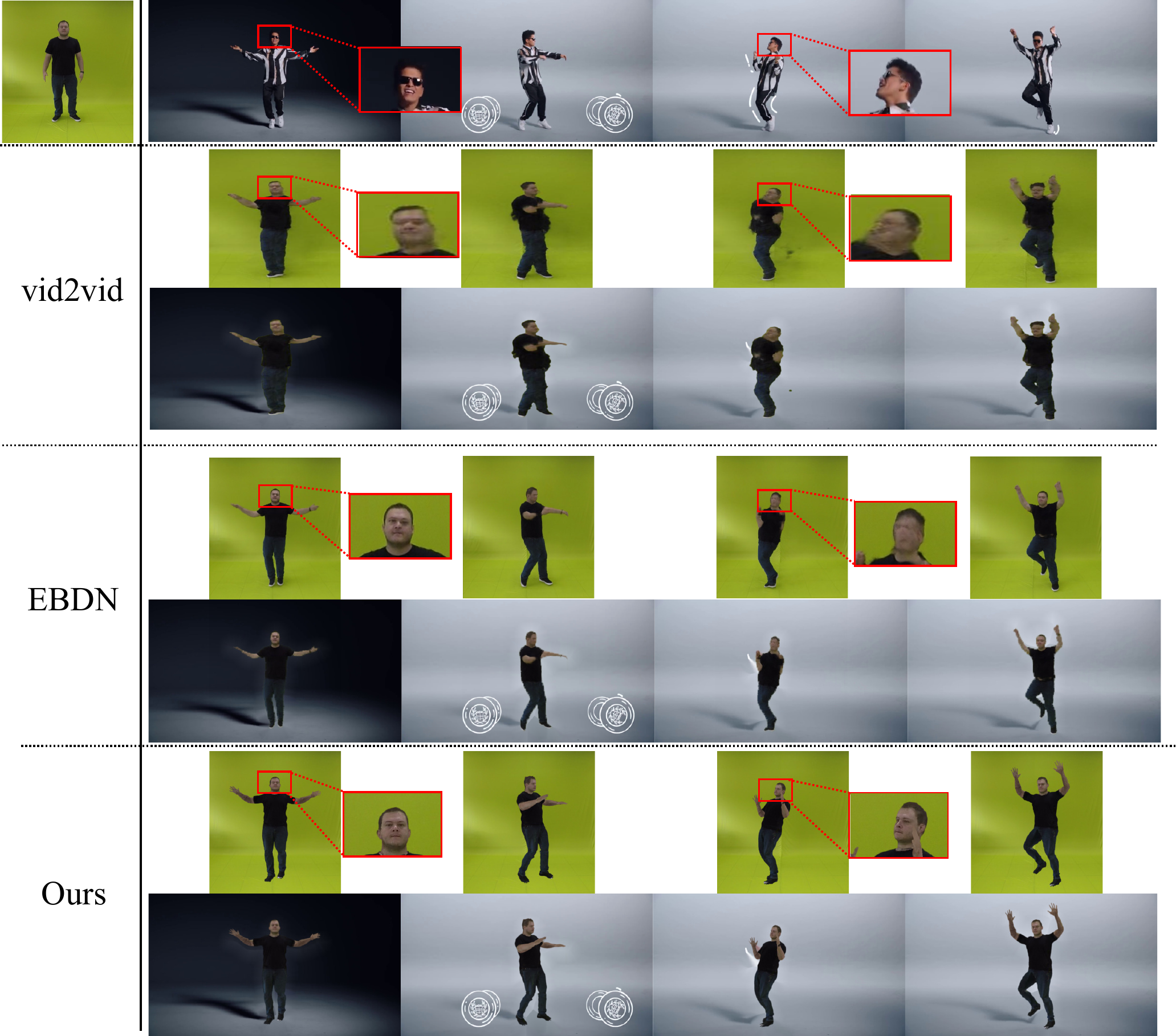}
	\caption{\textbf{Qualitative evaluation to bruno-mars sequence.} {\it First row:} original video and target actor S5; {\it Second row:} Result of vid2vid and their compositing with the target background; {\it Third row:} Results of EBDN and their compositing with the target background; Fourth row: Our results for both backgrounds.}
	\label{result:bruno}
\end{figure}

Some frames used to compute the metrics in Table~\ref{table:metrics_result} are shown in Figure~\ref{result:quali-dataset-0}. One can note the large discrepancy between the quality of the frames for the same method. As shown in Figures~\ref{result:quali-dataset-0} and~\ref{result:bruno}, the end-to-end learning techniques have impressive results when the person is standing parallel to the image plane and the arms are not in front of the face; however, these methods perform poorly when the person is out of these contexts, such as when bending. Our method, for its turn, retains the same quality for most poses. 

\begin{figure}[!t]
	\includegraphics[width=0.95\linewidth]{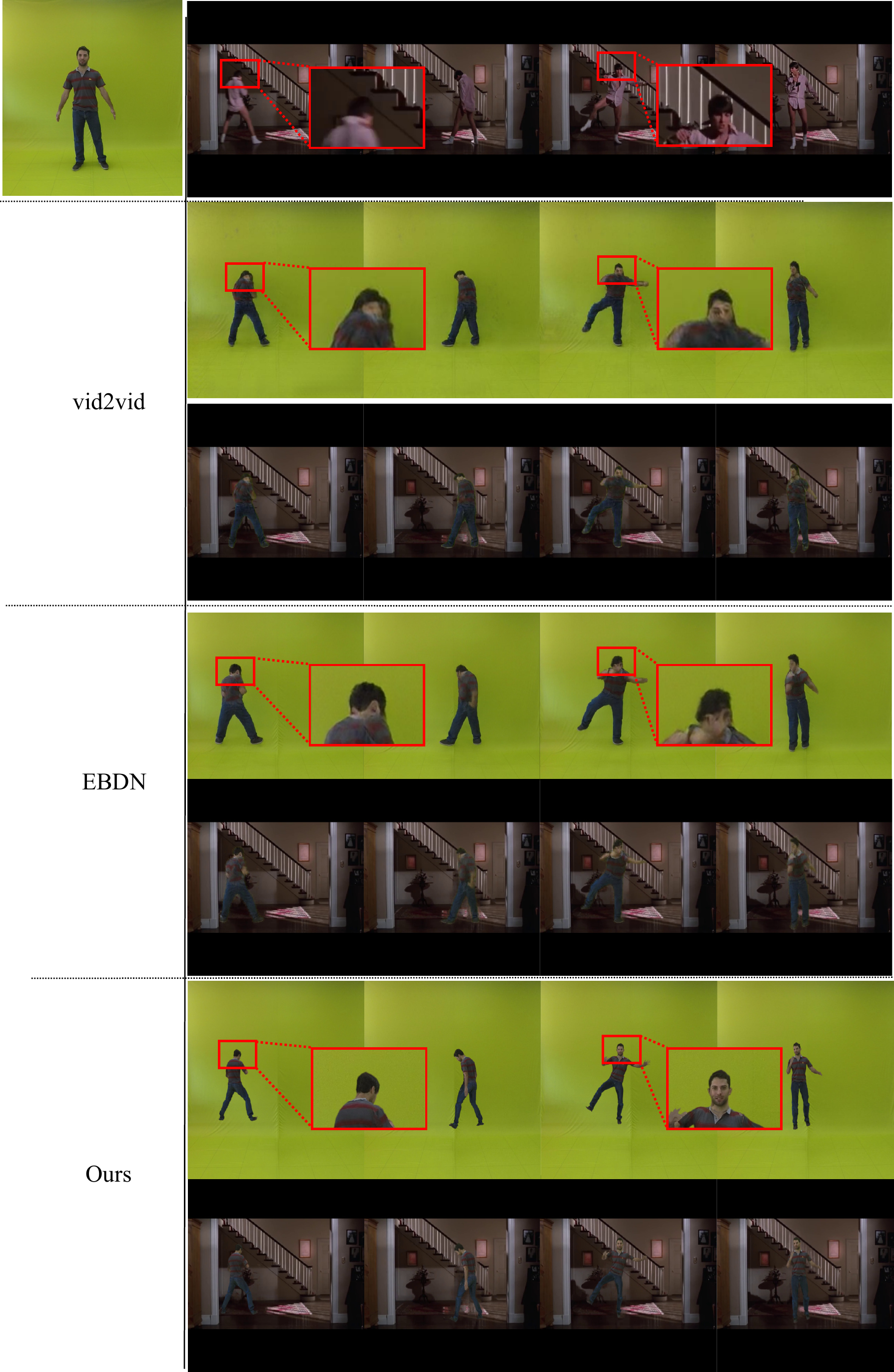}
	\caption{\textbf{Qualitative evaluation to tom-cruise sequence.} {\it First row:} original video and target actor S2; {\it Second row:} Result of vid2vid and their compositing with the target background; {\it Third row:} Result of EBDN and their compositing with the target background; {\it Fourth row:} Our results for both backgrounds.}
	\label{result:tom}
\end{figure}

We also analyzed the impact of the camera pose and the actor's scale in the quality of the resulting videos of the methods. We transferred two target persons from our dataset to two videos with different camera setups and image resolutions: ``bruno-mars'' and ``tom-cruise'' sequences. In the ``bruno-mars'' sequence, shown in Figure~\ref{result:bruno}, we ran vid2vid and EBDN using their respective solution to tackle with different image resolutions and actors with different proportions from the training data. Vid2vid's strategy (scaling to keep the aspect ratio and then cropping the image) results in an actor with different proportions compared with the training data, which leads to a degradation of quality. EBDN's strategy (pose normalization, scaling to keep the aspect ratio, and then cropping the image) keeps the similarity between the input video and training data, but body and face occlusions are still problems. The red squares highlight these issues on the right side in Figure~\ref{result:bruno}, where EBDN and vid2vid reconstructed poorly the target's face. These strategies also incur a loss of the relative position in the original image; thus, whenever the actor presents a large translation, he/she can stay out of the crop area. Figure~\ref{result:tom} depicts the results of ``tom-cruise'' sequence. In this sequence, the source actor has a large translation in all directions; thus, we include a margin in the image to allow EBDN and vid2vid to process it. Nevertheless, the difference in the actor’s scale results in low quality for vid2vid and EBDN, which suggests their lack of generalization of these methods to different poses, and changes in camera viewpoint and intrinsic parameters. On its turn, our method did not suffer from problems caused by the camera and image resolution, and provided the best scores and visual results.

\subsubsection{Results in the iPER Dataset}

Aside from our dataset, we also evaluated our approach using the iPER dataset~\citep{lwb2019}. Since the iPER dataset does not provide paired motions as our dataset, we evaluated our method according to~\cite{lwb2019}'s protocol, where SSIM and LPIPS are used to measure self-imitation transferring. We also include FVD in the evaluation, which was designed to capture the temporal coherence among videos and frame quality.

\begin{figure}[!t]
	\includegraphics[width=0.95\linewidth]{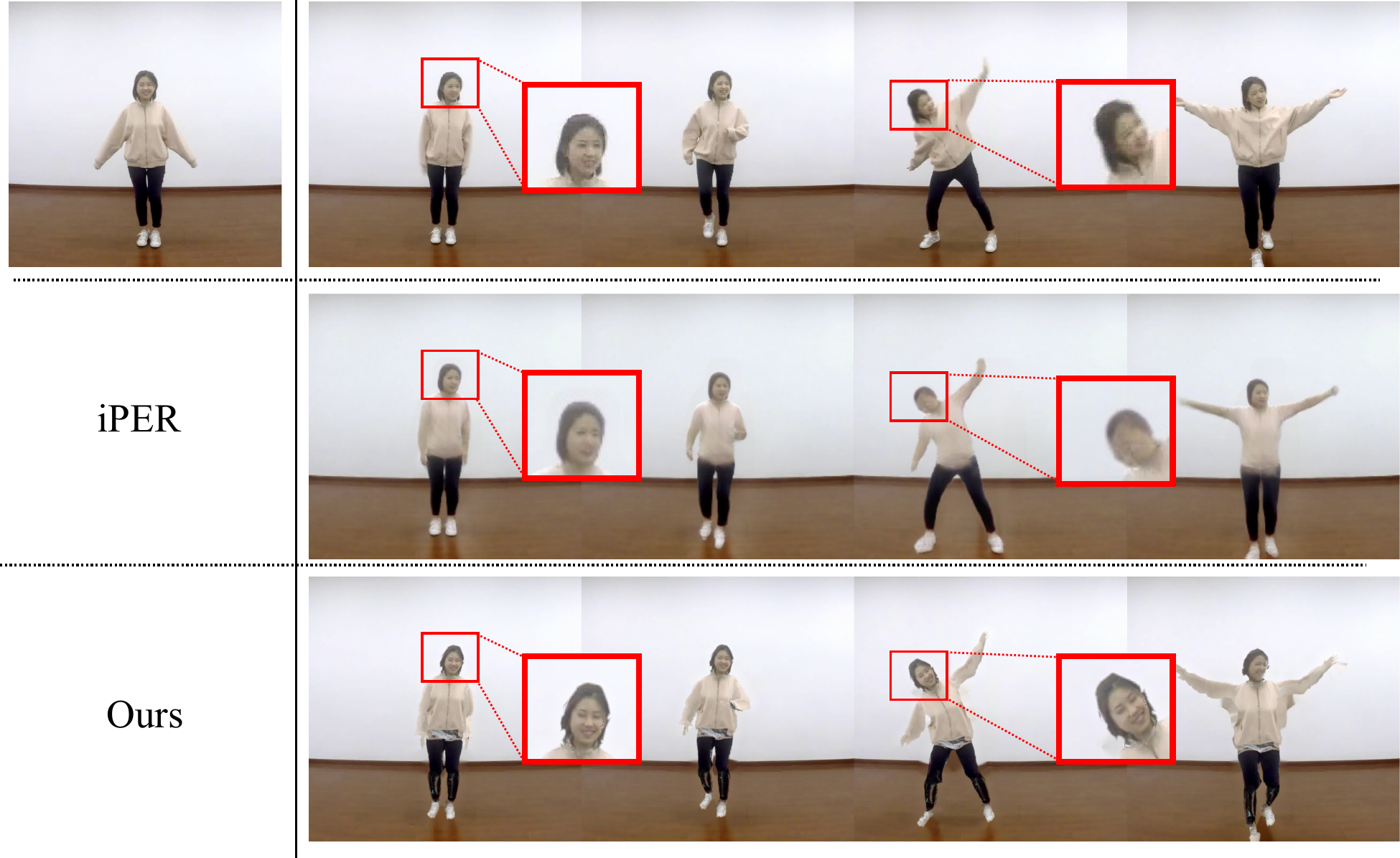}
	\caption{
		\textbf{Qualitative evaluation on a sequence from the iPER dataset.} {\it First row:} target actor and original video; {\it Second row:} Results of iPER network; {\it Third row:} Our results.}
	\label{result:iper}
\end{figure}

Table~\ref{table:iper_dataset} shows the transferring results in the iPER dataset. It can be seen that our method outperforms iPER in terms of SSIM and LPIPS metrics. This result also concurs with the visual results shown in Figure~\ref{result:iper}. We can also note that iPER achieved the best FVD value, since they are tested now in the same context where it was trained. However, it still performs poorly when the person is not standing up straight, as indicated in the facial zoom (red box in Figure~\ref{result:iper}). Our method retains the same quality for most poses.

\begin{table}[t!]
	\centering
	\caption{{\bf Comparison with iPER in their proposed dataset}. Our approach is able to provide the best values in terms of SSIM and LPIPS, while performing worse in terms of FVD (best in bold).}

	\label{table:iper_dataset}

	\begin{tabular}{@{}clcc@{}}
		\toprule 
		& {\bf Metric}& {\bf iPER} & {\bf Ours} \\ \cmidrule{2-4}

		& SSIM$\uparrow$   & $0.8410$ & $\mathbf{0.8936}$ \\
		& LPIPS$\downarrow$  & $0.0848$ & $\mathbf{0.0722}$ \\
		& FVD$\downarrow$  & $\mathbf{955}$ & $1269$ \\
		
		\bottomrule
	\end{tabular}
	
\end{table}

\subsection{Applications and Limitations}

\begin{figure}
	\includegraphics[width=0.95\linewidth]{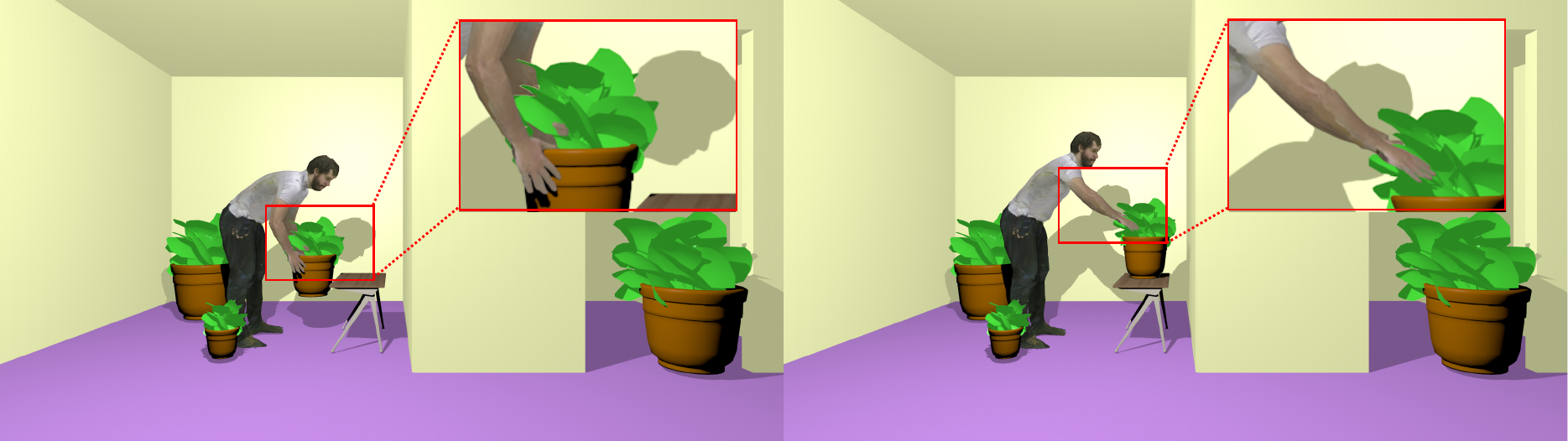}
	\caption{\textbf{Model-to-virtual}. Our method is able to provide a deformed model from images that can be included in virtual scenes using existing graphic rendering tools, such as Blender. The partial occlusions between the scene and model are handled in a natural way (red squares).}
	\label{result:model-to-virtual}
\end{figure}

In our experiments, we observed that image-based rendering and 3D reasoning still have several advantages regarding end-to-end learning human view synthesis, particularly in terms of control and generalizing to furthest views. Our method enables the replacement of the background, which is crucial for many applications. Moreover, it also allows to create a deformed model from images that can be included in virtual scenes using existing rendering frameworks, such as Blender. Figure~\ref{result:model-to-virtual} illustrates an application of this capability.

\begin{figure}
	\includegraphics[width=0.95\linewidth]{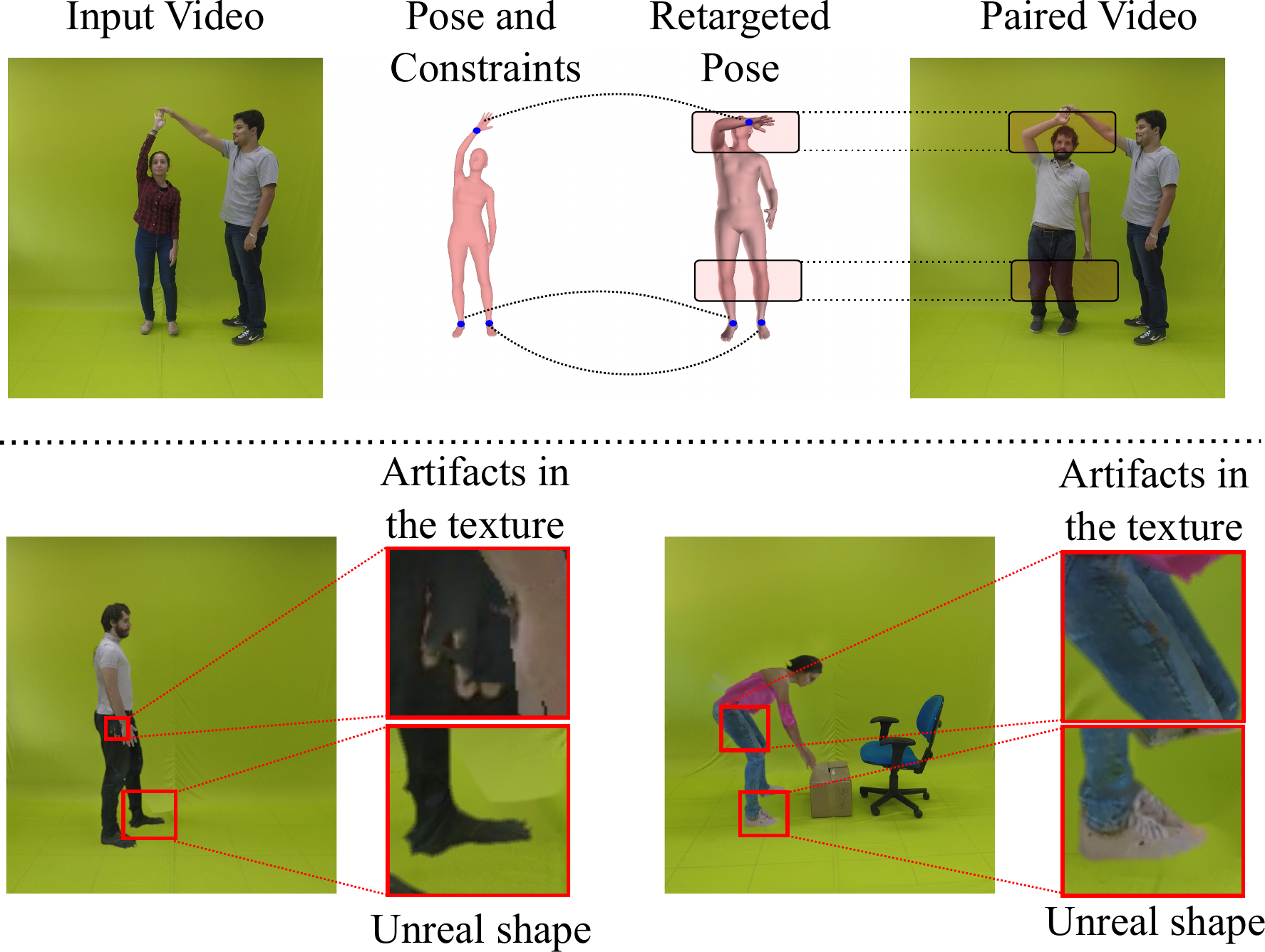}
	\caption{\textbf{Limitations}. {\it Top:} Retargeting resulting in undesired motion where the actor positions his hand down and curves his back instead of bending his knees. {\it Bottom:} Typical failure cases in the avatar: artifacts in the texture and body parts with unreal shapes.}
	\label{result:limitations}
\end{figure}

Although achieving the best results in these more generic test conditions, the proposed approach also suffers from certain limitations. Ideally, the constrained optimization problem would maintain the main features of the source motion. However, there is no single solution to a set of constraints, which can result in undesired motions, as shown in Figure~\ref{result:limitations}, where the actor positions his hand down and curve his back instead of bending his knees. The textured avatars may exhibit artifacts in the presence of part segmentation estimation errors, which can also lead to wrong deformations, and errors in the deformation result in body parts with unreal shapes. Typical failure cases such as retargeting and segmentation errors are also depicted in Figure~\ref{result:limitations}.

Regarding our method's processing time, our current implementation is a modular Python code, but without any optimization, \ie, the code was not parallelized either and was not adapted to run fully on a GPU. However, we highlight the available room for speeding up the processing time with a parallel implementation of different parts of the approach as, for instance, in the deformation model steps, which currently require $30$ seconds per frame. They could be easily adapted to be executed in parallel with multiprocessing.

Finally, in the proposed dataset, all videos were selected and designed to stress the evaluation of the human motion and appearance transfer capabilities of the methods for several paired movements of different actors. We adopted a simpler background to avoid eventual errors that could be added during the evaluation from the foreground-background segmentation (which is done before inpainting). This choice is meant to allow a better assessment of each method's capabilities to transfer actors' motions and appearance with different morphologies. However, there is still a lack of paired videos with complex backgrounds to assess the quality of the human retargeting and the background synthesis together.

\section{Conclusions}

This paper proposes a new model-based retargeting methodology that incorporates different strategies to extract 3D shape, pose, and appearance to transfer motions between two real human characters using the information from monocular videos. Unlike retargeting methods that use either appearance information or motion information only, our approach simultaneously considers the critical factors to retargeting such as pose, shape, appearance, and motion features. 

Furthermore, we have introduced a new dataset comprising different videos with annotated human-to-object interactions, actors with different body shapes, and paired motions to evaluate the task of retargeting humans in videos. We believe that these videos can be used as a common base to improve tracking the field's progress by showing where current approaches might fail and measuring the quality of future proposals. 

Despite the impressive advances in image-to-image translation approaches, we presented several contexts where most end-to-end image translation methods perform poorly. Yet, our experiments show that a model-based rendering technique can still exhibit a competitive quality compared to recent end-to-end learning techniques, besides having several advantages in terms of control and ability to generalize to furthest views. Our results indicate that retargeting strategies based on image-to-image learning are still challenged to retarget motions while keeping the desired movement constraints, shape, and appearance simultaneously. These findings suggest the potential of hybrid strategies by leveraging the advantages provided by model-based retargeting into recent neural rendering approaches. \\ 

\noindent\textbf{Acknowledgements.} The authors thank CAPES, CNPq, and FAPEMIG for funding this work. We also thank NVIDIA for the donation of a Titan XP GPU used in this research.





{ \small
	\bibliographystyle{spbasic}      
	\bibliography{references} 

\begin{thebibliography}{54}
\providecommand{\natexlab}[1]{#1}
\providecommand{\url}[1]{{#1}}
\providecommand{\urlprefix}{URL }
\expandafter\ifx\csname urlstyle\endcsname\relax
  \providecommand{\doi}[1]{DOI~\discretionary{}{}{}#1}\else
  \providecommand{\doi}{DOI~\discretionary{}{}{}\begingroup
  \urlstyle{rm}\Url}\fi
\providecommand{\eprint}[2][]{\url{#2}}

\bibitem[{Aberman et~al.(2018)Aberman, Shi, Liao, Lischinski, Chen, and
  Cohen{-}Or}]{Aberman_2018}
Aberman, K., Shi, M., Liao, J., Lischinski, D., Chen, B., Cohen{-}Or, D. (2018)
  Deep video-based performance cloning. {\em CoRR}

\bibitem[{Aberman et~al.(2019)Aberman, Wu, Lischinski, Chen, and
  Cohen-Or}]{retargeting_2d}
Aberman, K., Wu, R., Lischinski, D., Chen, B., Cohen-Or, D. (2019) Learning
  character-agnostic motion for motion retargeting in 2d. {\em ACM TOG}

\bibitem[{Alldieck et~al.(2018)Alldieck, Magnor, Xu, Theobalt, and
  Pons-Moll}]{alldieck2018video}
Alldieck, T., Magnor, M., Xu, W., Theobalt, C., Pons-Moll, G. (2018) Video
  based reconstruction of 3d people models. In: {\em CVPR}

\bibitem[{Andriluka et~al.(2014)Andriluka, Pishchulin, Gehler, and
  Schiele}]{MPII_dataset}
Andriluka, M., Pishchulin, L., Gehler, P., Schiele, B. (2014) 2d human pose
  estimation: New benchmark and state of the art analysis. In: {\em CVPR}

\bibitem[{Anguelov et~al.(2005)Anguelov, Srinivasan, Koller, Thrun, Rodgers,
  and Davis}]{Anguelov_2005}
Anguelov, D., Srinivasan, P., Koller, D., Thrun, S., Rodgers, J., Davis, J.
  (2005) Scape: Shape completion and animation of people. {\em ACM Trans Graph}

\bibitem[{Balakrishnan et~al.(2018)Balakrishnan, Zhao, Dalca, Durand, and
  Guttag}]{Balakrishnan}
Balakrishnan, G., Zhao, A., Dalca, A.V., Durand, F., Guttag, J.V. (2018)
  Synthesizing images of humans in unseen poses. In: {\em CVPR}

\bibitem[{Bau et~al.(2019)Bau, Zhu, Wulff, Peebles, Strobelt, Zhou, and
  Torralba}]{bau2019seeing}
Bau, D., Zhu, J.Y., Wulff, J., Peebles, W., Strobelt, H., Zhou, B., Torralba,
  A. (2019) Seeing what a gan cannot generate. In: {\em ICCV}

\bibitem[{Bogo et~al.(2016)Bogo, Kanazawa, Lassner, Gehler, Romero, and
  Black}]{Bogo_2016}
Bogo, F., Kanazawa, A., Lassner, C., Gehler, P., Romero, J., Black, M.J. (2016)
  Keep it smpl: Automatic estimation of 3d human pose and shape from a single
  image. In: {\em ECCV}

\bibitem[{Cao et~al.(2017)Cao, Simon, Wei, and Sheikh}]{openpose1}
Cao, Z., Simon, T., Wei, S.E., Sheikh, Y. (2017) Realtime multi-person 2d pose
  estimation using part affinity fields. In: {\em CVPR}

\bibitem[{{Chan} et~al.(2019){Chan}, {Ginosar}, {Zhou}, and
  {Efros}}]{chan2018dance}
{Chan}, C., {Ginosar}, S., {Zhou}, T., {Efros}, A. (2019) Everybody dance now.
  In: {\em ICCV}

\bibitem[{Choi and Ko(2000)}]{motion_retargetting_1}
Choi, K.J., Ko, H.S. (2000) On-line motion retargeting. {\em Journal of
  Visualization and Computer Animation}

\bibitem[{Criminisi et~al.(2004)Criminisi, Perez, and
  Toyama}]{based-inpainting}
Criminisi, A., Perez, P., Toyama, K. (2004) Region filling and object removal
  by exemplar-based image inpainting. {\em IEEE TIP}

\bibitem[{De~Boor et~al.(1978)De~Boor, De~Boor, Math{\'e}maticien, De~Boor, and
  De~Boor}]{splines}
De~Boor, C., De~Boor, C., Math{\'e}maticien, E.U., De~Boor, C., De~Boor, C.
  (1978) A practical guide to splines, vol~27. Springer Verlag

\bibitem[{Dosovitskiy et~al.(2015)Dosovitskiy, Springenberg, and
  Brox}]{CNN_for_view_synthesis_1}
Dosovitskiy, A., Springenberg, J.T., Brox, T. (2015) Learning to generate
  chairs with convolutional neural networks. In: {\em CVPR}

\bibitem[{Esser et~al.(2018)Esser, Sutter, and Ommer}]{Esser_2018_CVPR}
Esser, P., Sutter, E., Ommer, B. (2018) A variational u-net for conditional
  appearance and shape generation. In: {\em CVPR}

\bibitem[{Gleicher(1998)}]{Gleicher}
Gleicher, M. (1998) Retargetting motion to new characters. In: {\em SIGGRAPH}

\bibitem[{Gomes et~al.(2020)Gomes, Martins, Ferreira, and
  Nascimento}]{Gomes_2020_WACV}
Gomes, T., Martins, R., Ferreira, J., Nascimento, E. (2020) Do as i do:
  Transferring human motion and appearance between monocular videos with
  spatial and temporal constraints. In: {\em WACV}

\bibitem[{Gong et~al.(2018)Gong, Liang, Li, Chen, Yang, and
  Lin}]{Gong2018InstancelevelHP}
Gong, K., Liang, X., Li, Y., Chen, Y., Yang, M., Lin, L. (2018) Instance-level
  human parsing via part grouping network. In: {\em ECCV}

\bibitem[{Hassan et~al.(2019)Hassan, Choutas, Tzionas, and Black}]{Hassan2019}
Hassan, M., Choutas, V., Tzionas, D., Black, M.J. (2019) Resolving {3D} human
  pose ambiguities with {3D} scene constraints. In: {\em ICCV}

\bibitem[{Kanazawa et~al.(2018)Kanazawa, Black, Jacobs, and
  Malik}]{kanazawaHMR18}
Kanazawa, A., Black, M.J., Jacobs, D.W., Malik, J. (2018) End-to-end recovery
  of human shape and pose. In: {\em CVPR}

\bibitem[{Kolotouros et~al.(2019)Kolotouros, Pavlakos, Black, and
  Daniilidis}]{kolotouros2019spin}
Kolotouros, N., Pavlakos, G., Black, M.J., Daniilidis, K. (2019) Learning to
  reconstruct 3d human pose and shape via model-fitting in the loop. In: {\em
  ICCV}

\bibitem[{Lassner et~al.(2017{\natexlab{a}})Lassner, Pons-Moll, and
  Gehler}]{Lassner_GeneratingPeople}
Lassner, C., Pons-Moll, G., Gehler, P.V. (2017{\natexlab{a}}) A generative
  model for people in clothing. In: {\em ICCV}

\bibitem[{Lassner et~al.(2017{\natexlab{b}})Lassner, Romero, Kiefel, Bogo,
  Black, and Gehler}]{Lassner_2017}
Lassner, C., Romero, J., Kiefel, M., Bogo, F., Black, M.J., Gehler, P.V.
  (2017{\natexlab{b}}) Unite the people: Closing the loop between 3d and 2d
  human representations. In: {\em CVPR}

\bibitem[{Levi and Gotsman(2015)}]{Zohar2015}
Levi, Z., Gotsman, C. (2015) Smooth rotation enhanced as-rigid-as-possible mesh
  animation. {\em T-VCG}

\bibitem[{Lin et~al.(2014)Lin, Maire, Belongie, Hays, Perona, Ramanan,
  Doll{\'a}r, and Zitnick}]{COCO_dataset}
Lin, T.Y., Maire, M., Belongie, S., Hays, J., Perona, P., Ramanan, D.,
  Doll{\'a}r, P., Zitnick, C.L. (2014) Microsoft coco: Common objects in
  context. In: {\em ECCV}

\bibitem[{{Liu} et~al.(2019){Liu}, {Piao}, {Jie}, {Luo}, {Ma}, and
  {Gao}}]{lwb2019}
{Liu}, W., {Piao}, Z., {Jie}, M., {Luo}, W., {Ma}, L., {Gao}, S. (2019) Liquid
  warping {GAN}: A unified framework for human motion imitation, appearance
  transfer and novel view synthesis. In: {\em ICCV}

\bibitem[{Loper et~al.(2015)Loper, Mahmood, Romero, Pons-Moll, and
  Black}]{Loper_2015}
Loper, M., Mahmood, N., Romero, J., Pons-Moll, G., Black, M.J. (2015) Smpl: A
  skinned multi-person linear model. {\em ACM Trans Graph}

\bibitem[{Ma et~al.(2017)Ma, Jia, Sun, Schiele, Tuytelaars, and
  Van~Gool}]{ma2017pose}
Ma, L., Jia, X., Sun, Q., Schiele, B., Tuytelaars, T., Van~Gool, L. (2017) Pose
  guided person image generation. In: {\em NIPS}

\bibitem[{Mahmood et~al.(2019)Mahmood, Ghorbani, Troje, Pons-Moll, and
  Black}]{AMASS:ICCV:2019}
Mahmood, N., Ghorbani, N., Troje, N.F., Pons-Moll, G., Black, M.J. (2019)
  {AMASS}: Archive of motion capture as surface shapes. In: {\em ICCV}

\bibitem[{{Marra} et~al.(2020){Marra}, {Gragnaniello}, {Verdoliva}, and
  {Poggi}}]{marra2019e2e}
{Marra}, F., {Gragnaniello}, D., {Verdoliva}, L., {Poggi}, G. (2020) A
  full-image full-resolution end-to-end-trainable cnn framework for image
  forgery detection. {\em IEEE Access}

\bibitem[{Mehta et~al.(2017)Mehta, Rhodin, Casas, Fua, Sotnychenko, Xu, and
  Theobalt}]{Mehta_2017}
Mehta, D., Rhodin, H., Casas, D., Fua, P., Sotnychenko, O., Xu, W., Theobalt,
  C. (2017) Monocular 3d human pose estimation in the wild using improved cnn
  supervision. In: {\em 3DV}

\bibitem[{Mir et~al.(2020)Mir, Alldieck, and Pons-Moll}]{mir20}
Mir, A., Alldieck, T., Pons-Moll, G. (2020) Learning to transfer texture from
  clothing images to 3d humans. In: {\em CVPR}, {IEEE}

\bibitem[{Neverova et~al.(2018)Neverova, G{\"{u}}ler, and
  Kokkinos}]{NeverovaGK18}
Neverova, N., G{\"{u}}ler, R.A., Kokkinos, I. (2018) Dense pose transfer. In:
  {\em ECCV}

\bibitem[{Peng et~al.(2018)Peng, Kanazawa, Malik, Abbeel, and
  Levine}]{2018-TOG-SFV}
Peng, X.B., Kanazawa, A., Malik, J., Abbeel, P., Levine, S. (2018) Sfv:
  Reinforcement learning of physical skills from videos. {\em ACM Trans Graph}

\bibitem[{Shysheya et~al.(2019)Shysheya, Zakharov, Aliev, Bashirov, Burkov,
  Iskakov, Ivakhnenko, Malkov, Pasechnik, Ulyanov, Vakhitov, and
  Lempitsky}]{Shysheya_2019_CVPR}
Shysheya, A., Zakharov, E., Aliev, K.A., Bashirov, R., Burkov, E., Iskakov, K.,
  Ivakhnenko, A., Malkov, Y., Pasechnik, I., Ulyanov, D., Vakhitov, A.,
  Lempitsky, V. (2019) Textured neural avatars. In: {\em CVPR}

\bibitem[{Sigal et~al.(2007)Sigal, Balan, and Black}]{Sigal}
Sigal, L., Balan, A., Black, M.J. (2007) Combined discriminative and generative
  articulated pose and non-rigid shape estimation. In: {\em NIPS}

\bibitem[{Simon et~al.(2017)Simon, Joo, Matthews, and Sheikh}]{openpose2}
Simon, T., Joo, H., Matthews, I., Sheikh, Y. (2017) Hand keypoint detection in
  single images using multiview bootstrapping. In: {\em CVPR}

\bibitem[{Sun et~al.(2020)Sun, Fu, Jiang, Liu, Lai, Fu, and Gao}]{sun2020human}
Sun, Y.T., Fu, Q.C., Jiang, Y.R., Liu, Z., Lai, Y.K., Fu, H., Gao, L. (2020)
  Human motion transfer with 3d constraints and detail enhancement.
  \eprint{2003.13510}

\bibitem[{Tatarchenko et~al.(2015)Tatarchenko, Dosovitskiy, and
  Brox}]{CNN_for_view_synthesis_0}
Tatarchenko, M., Dosovitskiy, A., Brox, T. (2015) Single-view to multi-view:
  Reconstructing unseen views with a convolutional network. {\em CoRR}

\bibitem[{Tewari et~al.(2020)Tewari, Fried, Thies, Sitzmann, Lombardi,
  Sunkavalli, Martin-Brualla, Simon, Saragih, Niebner, Pandey, Fanello,
  Wetzstein, Zhu, Theobalt, Agrawala, Shechtman, Goldman, and
  Zollhofer}]{tewari2020cgf}
Tewari, A., Fried, O., Thies, J., Sitzmann, V., Lombardi, S., Sunkavalli, K.,
  Martin-Brualla, R., Simon, T., Saragih, J., Niebner, M., Pandey, R., Fanello,
  S., Wetzstein, G., Zhu, J.Y., Theobalt, C., Agrawala, M., Shechtman, E.,
  Goldman, D.B., Zollhofer, M. (2020) State of the art on neural rendering.
  {\em Computer Graphics Forum} 39(2):701--727, \doi{10.1111/cgf.14022}

\bibitem[{Unterthiner et~al.(2019)Unterthiner, van Steenkiste, Kurach,
  Marinier, Michalski, and Gelly}]{unterthiner2019accurate}
Unterthiner, T., van Steenkiste, S., Kurach, K., Marinier, R., Michalski, M.,
  Gelly, S. (2019) Towards accurate generative models of video: A new metric \&
  challenges. \eprint{1812.01717}

\bibitem[{Villegas et~al.(2018)Villegas, Yang, Ceylan, and
  Lee}]{Villegas_2018_CVPR}
Villegas, R., Yang, J., Ceylan, D., Lee, H. (2018) Neural kinematic networks
  for unsupervised motion retargetting. In: {\em CVPR}

\bibitem[{Wang et~al.(2019)Wang, Huang, Han, and Wang}]{wang2018videoinp}
Wang, C., Huang, H., Han, X., Wang, J. (2019) Video inpainting by jointly
  learning temporal structure and spatial details. In: {\em AAAI}

\bibitem[{{Wang} et~al.(2020){Wang}, {Wang}, {Zhang}, {Owens}, and
  {Efros}}]{Wang2019CNNgeneratedIA}
{Wang}, S., {Wang}, O., {Zhang}, R., {Owens}, A., {Efros}, A.A. (2020)
  Cnn-generated images are surprisingly easy to spot… for now. In: {\em CVPR}

\bibitem[{Wang et~al.(2020)Wang, Wang, Zhang, Owens, and Efros}]{wang2020cvpr}
Wang, S.Y., Wang, O., Zhang, R., Owens, A., Efros, A.A. (2020) Cnn-generated
  images are surprisingly easy to spot... for now. In: {\em CVPR}

\bibitem[{Wang et~al.(2018)Wang, Liu, Zhu, Liu, Tao, Kautz, and
  Catanzaro}]{wang2018vid2vid}
Wang, T.C., Liu, M.Y., Zhu, J.Y., Liu, G., Tao, A., Kautz, J., Catanzaro, B.
  (2018) Video-to-video synthesis. In: {\em NIPS}

\bibitem[{Wang et~al.(2004)Wang, Bovik, Sheikh, and
  Simoncelli}]{Wang04imagequality}
Wang, Z., Bovik, A.C., Sheikh, H.R., Simoncelli, E.P. (2004) Image quality
  assessment: From error visibility to structural similarity. {\em IEEE TIP}

\bibitem[{Wei et~al.(2016)Wei, Ramakrishna, Kanade, and Sheikh}]{openpose3}
Wei, S.E., Ramakrishna, V., Kanade, T., Sheikh, Y. (2016) Convolutional pose
  machines. In: {\em CVPR}

\bibitem[{Xu et~al.(2019)Xu, Li, Zhou, and Loy}]{Xu_2019_CVPR}
Xu, R., Li, X., Zhou, B., Loy, C.C. (2019) Deep flow-guided video inpainting.
  In: {\em CVPR}

\bibitem[{Yang et~al.(2015)Yang, Reed, Yang, and
  Lee}]{CNN_for_view_synthesis_2}
Yang, J., Reed, S., Yang, M.H., Lee, H. (2015) Weakly-supervised disentangling
  with recurrent transformations for 3d view synthesis. In: {\em NIPS}

\bibitem[{Yu et~al.(2018)Yu, Lin, Yang, Shen, Lu, and Huang}]{yu2018generative}
Yu, J., Lin, Z., Yang, J., Shen, X., Lu, X., Huang, T.S. (2018) Generative
  image inpainting with contextual attention. In: {\em CVPR}

\bibitem[{{Yu} et~al.(2019){Yu}, {Lin}, {Yang}, {Shen}, {Lu}, and
  {Huang}}]{yu2018free}
{Yu}, J., {Lin}, Z., {Yang}, J., {Shen}, X., {Lu}, X., {Huang}, T. (2019)
  Free-form image inpainting with gated convolution. In: {\em ICCV}

\bibitem[{Zhang et~al.(2018)Zhang, Isola, Efros, Shechtman, and
  Wang}]{zhang2018perceptual}
Zhang, R., Isola, P., Efros, A.A., Shechtman, E., Wang, O. (2018) The
  unreasonable effectiveness of deep features as a perceptual metric. In: {\em
  CVPR}

\bibitem[{Zhao et~al.(2017)Zhao, Wu, Cheng, Liu, and Feng}]{ZhaoWCLF17}
Zhao, B., Wu, X., Cheng, Z., Liu, H., Feng, J. (2017) Multi-view image
  generation from a single-view. {\em CoRR}

\end{thebibliography}
}

\end{document}